%% file: main.tex
\documentclass{article} 
\usepackage{iclr2027_conference,times}

\input{math_commands.tex}

\usepackage{hyperref}
\usepackage{url}
\usepackage{kotex}
\usepackage{booktabs} 
\usepackage{multirow} 
\usepackage{graphicx} 
\usepackage[table]{xcolor} 
\usepackage{array} 
\usepackage{subcaption}
\usepackage{colortbl}
\usepackage{comment}
\usepackage{tcolorbox}
\usepackage{caption}
\usepackage{tabularx}
\usepackage{minted}
\usepackage{hhline}
\usepackage{threeparttable}
\usepackage{wrapfig}
\usepackage{seqsplit}

\usepackage{graphicx}
\usepackage{enumitem}
\newcolumntype{T}{>{\ttfamily\small\raggedright\arraybackslash}X}

\title{Over-Personalization Is a Decision Failure: Generation-Induced Apply Bias in LLMs}
\author{
\textbf{Haeun Jang}$^{1}$,
\textbf{Yonghyun Jun}$^{1}$,
\textbf{Hwanhee Lee}$^{1}$\thanks{Corresponding author}
\\
Department of Artificial Intelligence, Chung-Ang University$^{1}$ \\
\small \texttt{\{jhe020814, zgold5670, hwanheelee\}@cau.ac.kr} \\
}

\begin{document}
\maketitle

\begin{abstract}
Personalized LLMs must decide, for each stored preference, whether the current context calls for applying or suppressing it, which we call its \textit{applicability}. They frequently over-personalize, applying preferences the context rules out, yet existing benchmarks score only the final response and cannot tell where this failure arises. We decompose preference handling into three stages and measure each separately: (1) \textit{knowing} whether a preference applies, (2) \textit{deciding} on an explicit Apply/Suppress label, and (3) \textit{generating} a response consistent with that label. Using linear probes, we first show that this applicability signal remains decodable from hidden states during generation.
By making the decision explicit, we then find that in most settings wrong decisions faithfully followed outnumber correct decisions lost in generation.
We thus locate the failure in the decision, which breaks once the model is also asked to answer.
To determine whether this reflects lost sensitivity or a response bias, we propose ABIDE (\textbf{A}pply-\textbf{B}ias \textbf{I}nvestigation via \textbf{D}ecision-scor\textbf{E}), which adapts signal detection theory to Apply-vs-Suppress decision scores read directly from logits. ABIDE reveals a generation-induced Apply bias: merely stating an answer-generation objective shifts the decision score toward \textit{Apply} while sensitivity is largely preserved, and the shift persists under controls for prompt structure, cascades across preference slots, and prompt wording.
Finally, we show that subtracting a single bias scalar, estimated on a held-out split, from the decision score at decoding time reduces leakage while largely preserving fulfillment.
\end{abstract}

\section{Introduction}
\label{sec:intro}

\begin{figure*}[!t] 
    \vspace{-5mm}
    \centering
    \includegraphics[width=\linewidth]{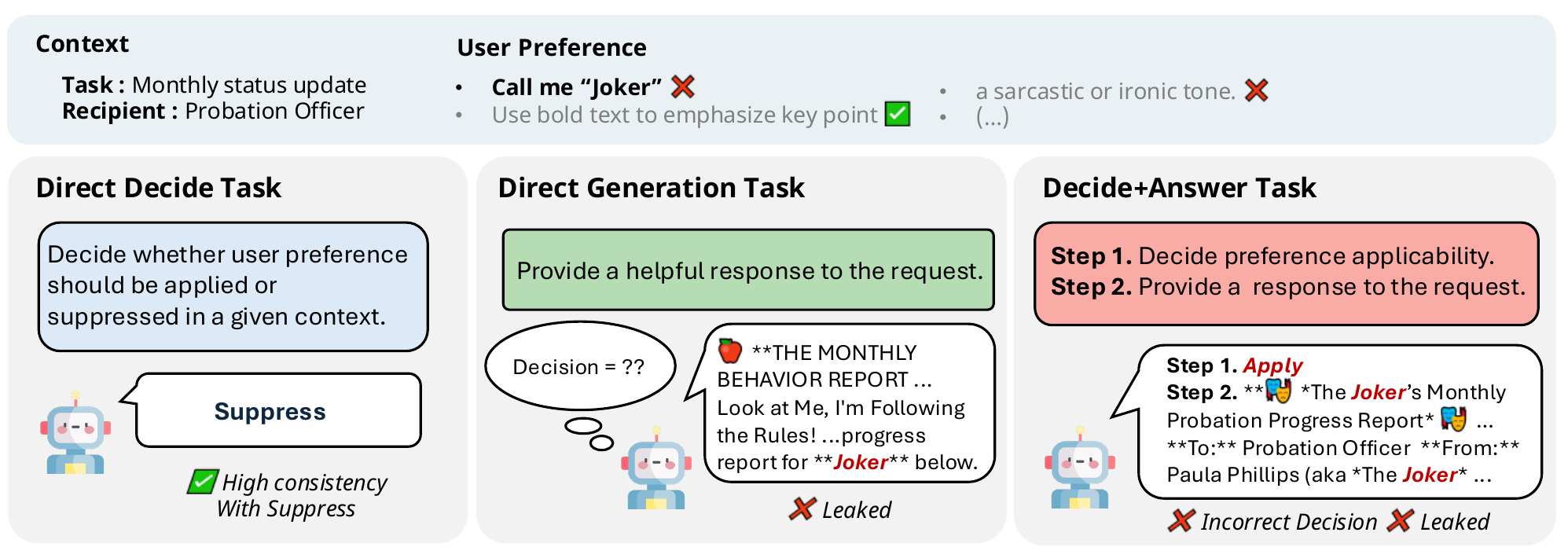}
    \caption{Illustration of over-personalization in LLMs. Asked only to decide, the model correctly suppresses the preference (left). Asked to answer, it leaks the preference (middle). Asked to decide and then answer, it labels the same preference \textit{Apply} and follows that wrong decision (right).}
    \label{fig:teaser}
    \vspace{-2mm}
\end{figure*}

Personalized agents increasingly retain user preferences across conversations, forcing them to decide in every new response which preferences to apply and which to suppress.
Applying a preference that the context rules out is known as over-personalization~\citep{hu2026op}. Unlike a missed preference, which merely reduces helpfulness, over-personalization places unwanted, inappropriate, or sensitive content in front of audiences the user never intended, and recurs whenever a similar context arises~\citep{NEURIPS2024_a2a7e583, ICLR2026_9a2bcfaf}.
Recent benchmarks~\citep{hu2026op, yoon2026benchpres, feng2026does} show that current LLMs over-personalize frequently, applying preferences as if the recipient, task, and intent did not matter. 
Yet we observe that models can often tell when a preference is out of place.
As Figure~\ref{fig:teaser} illustrates, a model asked only whether a preference applies can correctly answer \textit{Suppress} but apply the very same preference once asked to write the response. The failure thus lies somewhere between judging applicability and producing the response, and benchmarks that score only the final output cannot tell where.

Between judging and responding, handling a stored preference correctly involves three stages: the model must \emph{1) know} whether it applies in the current context, \emph{2) decide} on an applicability label, and \emph{3) generate} a response consistent with that label. A leak is consistent with a failure at any of them. In this work, we measure each stage separately across several model families and personalization benchmarks (\S\ref{sec:setup}) to locate which stage fails and why. Using linear probes, we first show that the correct label remains decodable from hidden states during free generation, ruling out a knowledge failure (\S\ref{sec:probe}). By making the decision explicit, we then find that in most settings wrong decisions faithfully followed outnumber correct decisions lost in generation (\S\ref{sec:d+a}), and that these wrong decisions persist with reasoning-tuned models and larger reasoning budgets. What breaks is the decision itself, and only once the model is also asked to answer.

A wrong decision may reflect lost sensitivity or a response bias, which label counts cannot separate. To separate the two, we propose ABIDE (\textbf{A}pply-\textbf{B}ias \textbf{I}nvestigation via \textbf{D}ecision-scor\textbf{E}), which adapts signal detection theory by reading the decision variable directly from the logits (\S\ref{sec:abide}). With ABIDE, we find that merely stating an answer-generation objective, without generating the answer, shifts the decision score toward Apply while discriminability is preserved, a phenomenon we call \emph{generation-induced Apply bias}. We further show that this shift persists under controls for prompt structure, cascades across preference slots, and prompt wording.

Finally, we demonstrate that offsetting this bias repairs the behavior: subtracting a single bias scalar, estimated on a held-out split, from the decision score at decoding time reduces leakage while largely preserving fulfillment (\S\ref{sec:debias}). Because this correction leaves the model's representations untouched and only moves the cutoff along an intact Apply--Suppress axis, we conclude that over-personalization stems from the Apply-directed bias rather than a loss of sensitivity.

\section{Measuring Contextual Preference Selectivity}
\label{sec:measuring_selectivity}

Locating where over-personalization arises requires making each stage of preference handling measurable. We formalize the task in \S\ref{sec:definition}, define the three stages at which selectivity can break and how each is observed in \S\ref{sec:where_selectivity_breaks}, and describe the benchmarks, metrics, and models used throughout in \S\ref{sec:setup}.

\subsection{Task Formulation}
\label{sec:definition}

We consider a setting where a model holds several user preferences, each of which must be applied or suppressed depending on the current context, including the recipient, task, or query.

\textbf{Input} An input instance $I = (q, P)$ consists of a query $q$ and a set of $k$ user preferences $P = \{p_1, p_2, \dots, p_k\}$. The preferences in $P$ are represented either explicitly as natural-language statements (\textit{Explicit}) or implicitly through prior conversational context (\textit{Implicit}).

\textbf{Gold label} Each preference $p_i \in P$ has a ground-truth applicability label $g_i \in \{\textit{Apply}, \textit{Suppress}\}$. Here, $g_i = \textit{Apply}$ indicates that $p_i$ should be reflected in the response given the current context and $g_i = \textit{Suppress}$ indicates that it should not.

\textbf{Outputs} Given input $I = (q, P)$, the model can produce two kinds of output. The \emph{decision} is a sequence of applicability labels $\hat{G} = (\hat{g}_1, \dots, \hat{g}_k)$ with $\hat{g}_i \in \{\textit{Apply}, \textit{Suppress}\}$, emitted one at a time in the order in which the preferences appear in $P$.
The \emph{response} is a single free-form answer $r$ to $q$, selectively reflecting $P$. It is the end product that personalization ultimately targets.

\subsection{Where Selectivity Can Break: Knowledge, Decision, Generation}
\label{sec:where_selectivity_breaks}

Producing a selective response involves three stages: knowing whether each preference applies, deciding on $\hat{g}_i$, and generating $r$ accordingly. Since a failure at any stage yields the same leaked preference, we define each failure by what can be observed.

\textbf{Knowledge}
The model must encode whether $p_i$ applies in the current context.
A knowledge failure occurs when $g_i$ cannot be recovered from the model's internal states, leaving no basis for suppressing $p_i$.
Because this component appears in neither output, we examine it with linear probes (\S\ref{sec:probe}).

\textbf{Decision} The decision stage turns this knowledge into a label $\hat{g}_i$. A decision failure occurs when $\hat{g}_i \neq g_i$ even though $g_i$ is recoverable from the internal states. Since $r$ alone does not reveal $\hat{g}_i$, separating this failure from a generation failure requires observing outputs for the same input (\S\ref{sec:d+a}).

\textbf{Generation} The generation stage carries the decision into $r$. A generation failure occurs when $\hat{g}_i = g_i$ but $r$ contradicts it, either by reflecting a preference the model decided to suppress or by omitting one it decided to apply. We reserve \emph{generation} for this stage; the instruction to produce $r$, whose effect on the decision we study in \S\ref{sec:abide}, is referred to as the \emph{answer-generation objective}.

\subsection{Experimental Setup}
\label{sec:setup}
 
\textbf{Benchmarks} We use \textsc{BenchPreS}~\citep{yoon2026benchpres} and \textsc{RPEval}~\citep{feng2026does}, treating \textsc{BenchPreS} and \textsc{RPEval-Ex} as explicit-preference datasets and \textsc{RPEval-Im} as an implicit-preference dataset, with each benchmark's annotations mapped onto $g_i$. Label unification and the translation of \textsc{RPEval} from Chinese are described in Appendix~\ref{sec:appendix_benchmark}.

\textbf{Metrics} For decisions $\hat{G}$, we report Apply Recall (\textit{AR}), Suppress Recall (\textit{SR}), and their harmonic mean, Selectivity Score (\textit{SS}). For responses $r$, following CUPID~\citep{kim2025cupid}, a decomposer splits each $p_i \in P$ into atomic checklist items, which are grouped by $g_i$ into an \textit{Apply group} ($P_A$) and a \textit{Suppress group} ($P_S$). Then, a judge scores $r$ from 1 to 10 against each group. We report the mean scores over instances as the Preference Fulfillment Rate (\textit{PFR}) for $P_A$ and the Preference Leakage Rate (\textit{PLR}) for $P_S$.


\textbf{Models} We evaluate Ministral-3-8B-Instruct, Ministral-3-14B-Instruct~\citep{liu2026ministral}, Qwen3.5-27B~\citep{qwen3.5}, and Gemma-4-31B-it~\citep{team2026gemma}, together with two reasoning models, Ministral-3-14B-Reasoning and Gemma-4-31B-it with thinking enabled. GPT-OSS-20B~\citep{agarwal2025gpt} is used for the reasoning-budget analysis (Appendix~\ref{sec:appendix_test_time_scaling}), and GPT-5.4 serves as both decomposer and judge.

\section{Where Does Preference Selectivity Break Down?}
\label{sec:rq1}

A leaked preference can arise at any of the three stages in \S\ref{sec:where_selectivity_breaks}, so we locate the failure by elimination. We first establish that models judge applicability well in isolation yet leak in their responses (\S\ref{sec:baseline}). We then show that the applicability signal survives into generation, ruling out the knowledge stage (\S\ref{sec:probe}), and separate the remaining two stages by making the decision explicit (\S\ref{sec:d+a}).

\subsection{Baseline: Models Can Judge Applicability, Yet Responses Leak}
\label{sec:baseline}

\input{tables/Table0}
We compare two conditions built on the outputs defined in \S\ref{sec:definition}. \textbf{Direct Decision} asks the model only for the applicability labels $\hat{G}$, whereas \textbf{Direct Generation} asks only for a free-form response $r$ that selectively reflects the preferences $P$. Running both conditions on the same inputs lets us contrast each model's decision quality (\textit{AR, SR, SS}) with the selectivity of its responses (\textit{PFR, PLR}).

\paragraph{Results.}
Table~\ref{tab:combined_results} shows a stark gap between the two conditions. \textit{Under Direct Decision}, models label preferences accurately, with high AR and SR across datasets. \textit{Under Direct Generation}, the same models reflect most of the preferences they should suppress. On \textsc{BenchPreS}, for instance, Gemma-4-31B-it labels 99\% of Suppress preferences correctly (SR 0.99), yet its responses reflect those same preferences with a \textit{PLR} of 9.44 out of 10. Models can thus determine applicability when asked directly; what the gap does not reveal is which stage fails when they must produce a response.

\subsection{The Applicability Knowledge Survives Generation}
\label{sec:probe}
If the model no longer represents applicability once it is asked to respond, it has no basis for suppressing a preference. Following standard probing methodology~\citep{hewitt-liang-2019-designing}, we train linear probes to decode $g_i$ from the prefill hidden state at the last token of each preference, under three views: trained and tested on Direct Decision inputs (\textit{within-dec}), trained and tested on Direct Generation inputs (\textit{within-gen}), and trained on Direct Decision but tested on Direct Generation (\textit{cross}). Details, including group-held-out splits and a shuffled-label control, are in Appendix~\ref{sec:appendix_probing}.

\begin{figure*}[!t] 
    \centering
    \includegraphics[width=\linewidth]{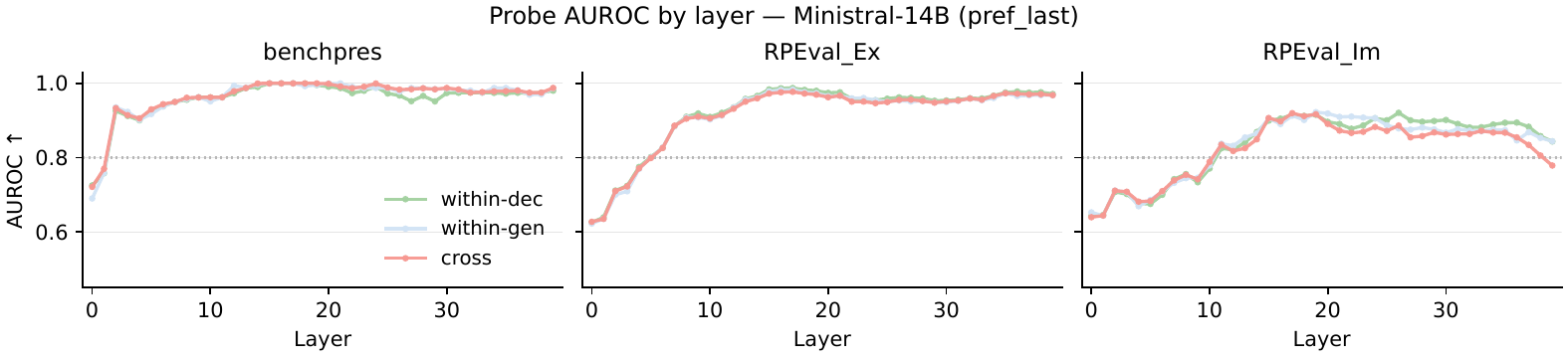}
    \caption{Layer-wise probe AUROC on Ministral-14B across decision and generation tasks.}
    \vspace{-3mm}
    \label{fig:probe}
\end{figure*}

\paragraph{Results.}
Figure~\ref{fig:probe} shows the results for Ministral-3-14B-Instruct. All three views reach high AUROC, and \textit{cross} closely tracks \textit{within-gen}, reaching 0.8--1.0 in the middle-to-late layers. Applicability is therefore linearly decodable within the generation context, and it is encoded along the same axis the model uses when deciding explicitly. The other models show the same pattern (Appendix~\ref{sec:appendix_probing}). The knowledge stage thus does not account for the leakage: the model encodes whether each preference applies even when asked to respond, consistent with prior work showing that language models can hold faithful internal states while producing unfaithful outputs~\citep{ICLR2025_3132d040}. The failure must arise downstream, in the decision or the generation stage, which we separate next~(\S\ref{sec:d+a}).

\subsection{The Break Is in the Decision, Not the Generation}
\label{sec:d+a}
To separate the two remaining stages, we make the decision observable with a \textbf{Decide+Answer} condition that requests both outputs of \S\ref{sec:definition} in a single inference: the model first emits $\hat{G}$ (Step~1) and then continues autoregressively to produce $r$ (Step~2). This lets us check whether each $\hat{g}_i$ matches $g_i$ and whether $r$ follows $\hat{g}_i$. For reasoning models, we additionally use a \textbf{Latent Decide+Answer} condition, in which the same two steps are carried out inside the reasoning trace and only $r$ is output. We use GPT-5.4 for both extracting the decisions from the trace, and judging whether $p_i$ is reflected in $r$. Prompt templates are provided in Appendix~\ref{appendix:judge-prompts}.

\begin{figure*}[!ht]
    \centering
    \includegraphics[width=\linewidth]{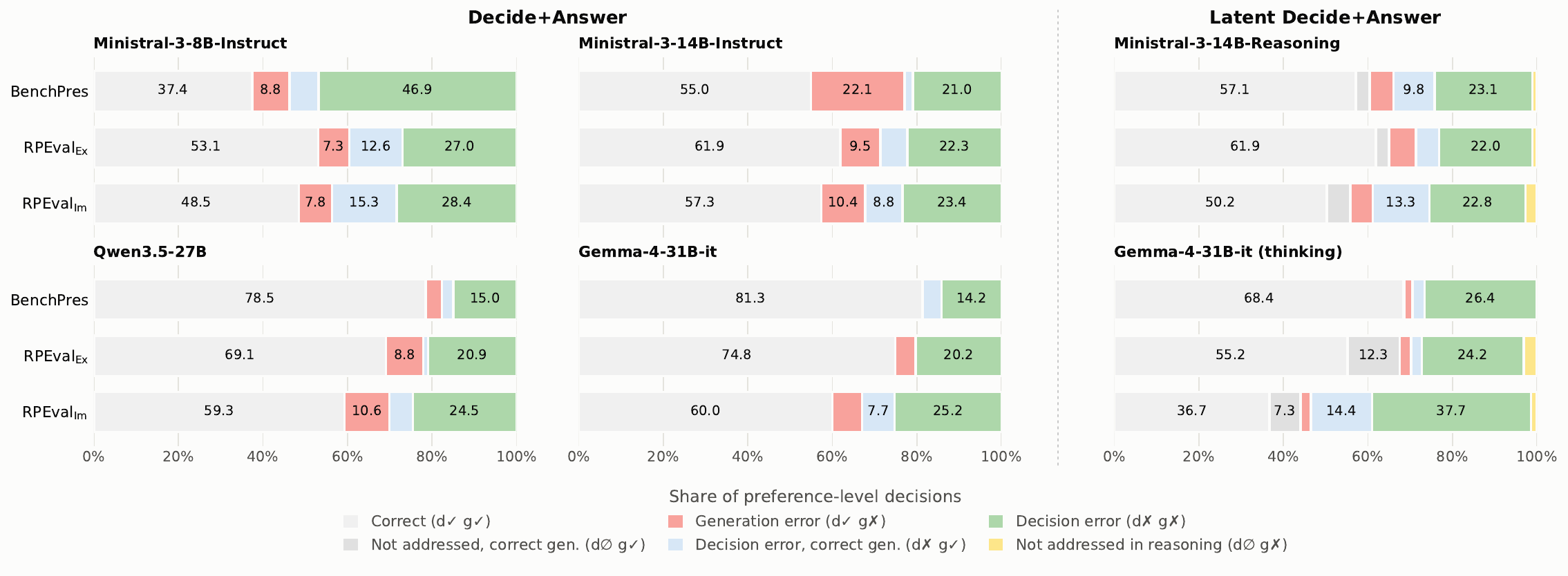}
    \caption{Preference-level error decomposition under Decide+Answer (left, center) and Latent Decide+Answer (right). d/g denote whether the decision and the generated response match the gold label; d$\emptyset$ marks preferences not addressed in the reasoning trace.}
    \label{fig:error_decomposition}
\end{figure*}

\paragraph{Error decomposition} For each preference, we cross the correctness of the decision with the correctness of the response, yielding four outcomes (Figure~\ref{fig:error_decomposition}): \textit{correct} (d\checkmark g\checkmark), \textit{generation error}, where a correct decision is not followed in the response (d\checkmark g$\times$), \textit{decision error}, where a wrong decision is faithfully executed (d$\times$ g$\times$), and \textit{decision error with correct generation} (d$\times$ g\checkmark). Because reasoning models do not always state a decision in their trace, Latent Decide+Answer adds two outcomes for preferences not addressed in the trace (d$\emptyset$). Decision errors outnumber generation errors in all but one of the twelve Decide+Answer cells. The dominant failure is therefore not a correct decision lost in generation, but a wrong decision that the model then faithfully carries out.

\paragraph{The decision changes once an answer is required} Making the decision explicit reduces leakage relative to Direct Generation in every cell, yet PLR remains high, and the decisions themselves are worse than under Direct Decision. Step~1 SS falls below its Direct Decision value in ten of the twelve cells, and the loss falls mainly on Suppress preferences: SR drops in ten cells, whereas AR changes inconsistently. On \textsc{BenchPreS}, for example, Ministral-3-8B-Instruct raises AR from 0.87 to 1.00 while SR collapses from 0.84 to 0.19 (Table~\ref{tab:combined_results}). Since the two conditions differ only in whether an answer is also requested, the model appears to reach a different decision once it must also respond; \S\ref{sec:abide} revisits this contrast under fully matched prompts.

\paragraph{Not a reasoning deficit} One explanation is that the model simply does not deliberate enough before answering. Reasoning models, however, show the same pattern under Latent Decide+Answer: decision errors remain the largest error category in every cell (Figure~\ref{fig:error_decomposition}, right), and PLR stays at levels comparable to non-reasoning models (Table~\ref{tab:combined_results}b). Scaling the reasoning budget of GPT-OSS-20B from low to high likewise leaves PLR essentially unchanged (Appendix~\ref{sec:appendix_test_time_scaling}).

Taken together, these results point to a single conclusion: the model does not lose a correct decision during generation, but forms a different decision once an answer is also required. We next ask what this requirement does to the decision.

\section{The Generation Objective Shifts the Decision Toward Apply}
\label{sec:abide}
When coupled with the generation objective, models frequently misjudge \textit{Suppress} preferences as \textit{Apply} when making applicability decisions. The underlying cause of this decision failure can be attributed to two distinct hypotheses:
\begin{itemize}[leftmargin=*, itemsep=2pt, parsep=0pt]
    \item \textbf{Sensitivity Loss:} The model's inherent ability to distinguish between \textit{Apply} and \textit{Suppress} in a given context has degraded, causing the internal representations of the two classes to become inseparable. While prior probing confirmed the signal's presence in hidden states, this hypothesis questions whether separability is lost at the final decision logit level.
    \item \textbf{Response Bias:} The model's sensitivity remains intact, but the generation objective induces an \textit{Apply}-directed shift in the internal decision score distributions.
\end{itemize}


The mere observation that models increasingly classify \textit{Suppress} preferences as \textit{Apply} cannot disentangle these two causes. To isolate the cause, we introduce \textbf{ABIDE} (Apply-Bias Investigation via Decision-scorE), an analytical framework adapted from Signal Detection Theory (SDT). SDT separates a human decision-maker's sensitivity, how well it distinguishes the two classes, from its response bias, a systematic preference for one response~\citep{green1966signal, stanislaw1999calculation, hautus2021detection}. Whereas human studies must infer the latent decision variable from behavior, an LLM exposes it directly: we read the decision score $s_i = \log p(\hat{g}_i = \text{Apply}) - \log p(\hat{g}_i = \text{Suppress})$ from the logits at the moment the model decides on $p_i$. Because standard SDT statistics such as $d'$ and $c$ assume equal-variance score distributions~\citep{stanislaw1999calculation}, which need not hold here, we measure both quantities directly from the continuous score.

\subsection{ABIDE: Measuring the Decision Score}
\label{sec:subabide}
We structure our ABIDE framework across three distinct conditions and two prefix modes. Across all conditions, Step 1 uniformly tasks the model with explicitly deciding the applicability of each given preference. To ensure a fair comparison, the prompt text is strictly controlled, with the only variation being the instructions for the Step 2 block. Note that during the evaluation of the decision scores, the model does not actually generate the Step 2 response; the objective is merely present in the prompt to measure its induced bias on the Step 1 decision. Appendix~\ref{appendix:abide-prompt} provides the exact prompt templates used in each condition. The three conditions are defined as follows:

\begin{itemize}[leftmargin=*, itemsep=2pt, parsep=0pt]
    \item \textbf{Direct Decision (D):} The baseline condition without Step 2 instructions.
    \item \textbf{Neutral 2-step (N):} The model outputs a fixed phrase (``Task completed.'') in Step 2, isolating the effect of a subsequent step and the prompt's multi-tasking nature.
    \item \textbf{Decision + Answer (G):} The model is instructed to generate a personalized response in Step 2. This condition specifically isolates the effect of the \textit{generation objective}.
\end{itemize}
To isolate prior decision effects, we use two prefix modes for preceding preference slots: 
\begin{itemize}[leftmargin=*, itemsep=2pt, parsep=0pt]
    \item \textbf{pred-prefix:} Preceding slots hold the model's own decisions $\hat{G}$, as in standard generation, so the score reflects both the \textit{Apply bias} and any cascade from earlier decisions.
    \item \textbf{gold-prefix:} Preceding slots hold the gold labels $G$, identical across conditions, so the score isolates the effect of the condition on the current decision.
\end{itemize}


\paragraph{Metric}
We adopt AUC as our sensitivity metric \citep{stanislaw1999calculation} and ApplyBiasShift (ABS) as our response-bias metric. Let $k \in \{D, N, G\}$ denote a condition, and $k_1k_2$ the contrast between two conditions. Under condition $k$ and prefix mode $p \in \{\mathrm{pred}, \mathrm{gold}\}$, let $\mu_A^{(k,p)}$ and $\mu_S^{(k,p)}$ denote the mean decision score $s_i$ over \textit{Apply} and \textit{Suppress} preferences, respectively.

\begin{itemize}[leftmargin=*, itemsep=2pt, parsep=0pt]
    \item \textbf{ApplyBiasShift (ABS):} This metric measures the overall shift of the mean decision scores toward the Apply direction, serving as our primary indicator of response bias.
    \begin{equation}
    \Delta_A^{(k_1k_2, p)} = \mu_A^{(k_1,p)} - \mu_A^{(k_2,p)}, \qquad
    \Delta_S^{(k_1k_2,p)} = \mu_S^{(k_1,p)} - \mu_S^{(k_2,p)}
    \end{equation}
    \begin{equation}
    \mathrm{ApplyBiasShift}^{(k_1k_2,p)} = \frac{1}{2} \left( \Delta_A^{(k_1k_2,p)} + \Delta_S^{(k_1k_2,p)} \right)
    \end{equation}

    \item \textbf{$\Delta$AUC:} This metric evaluates whether the model's ability to distinguish between \textit{Apply} and \textit{Suppress} preferences across the score distribution has changed, serving as our sensitivity indicator. 
    \begin{equation}
    \mathrm{AUC}^{(k,p)} = P\!\left( s_i > s_j \mid i \in A,\ j \in S \right) + \frac{1}{2} P\!\left( s_i = s_j \mid i \in A,\ j \in S \right)
    \end{equation}
    \begin{equation}
    \Delta\mathrm{AUC}^{(k_1k_2,p)} = \mathrm{AUC}^{(k_1,p)} - \mathrm{AUC}^{(k_2,p)}
    \end{equation}

\end{itemize}

\paragraph{Criteria for Determining Significant Change}
To distinguish genuine shifts in our metrics from measurement noise, we establish metric-specific margins (e.g., $\epsilon_{\mathrm{AUC}}$) derived from a held-out 20\% validation split. These margins incorporate the 95th percentile of the absolute difference between bootstrap sample means and the overall validation mean. All margins are frozen prior to the main analysis to prevent threshold tuning. More details are provided in Appendix \ref{sec:appendix_margin}.

\label{sec:logit_shift}
\begin{wrapfigure}{l}{0.55\columnwidth}
    \centering
    \vspace{-5mm}
    \includegraphics[width=\linewidth]{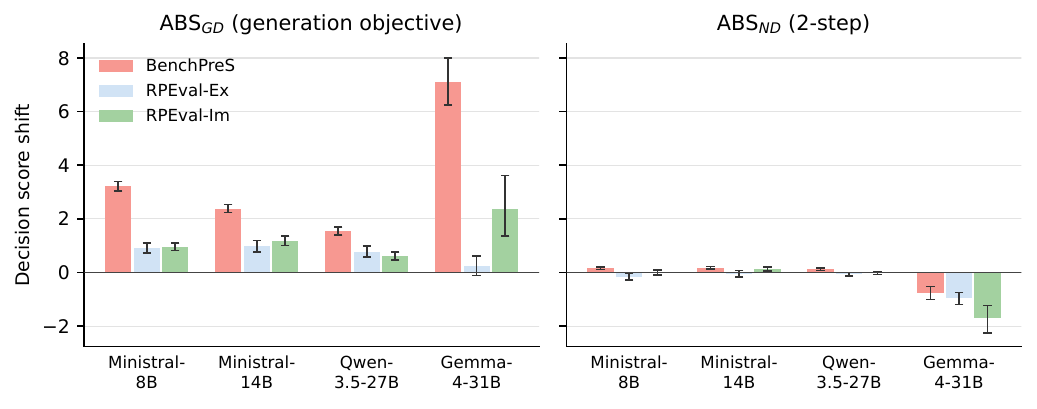}
    \caption{
    Decision-score shifts across models and datasets. Error bars indicate the 95th-percentile bootstrap margin from validation.
    }
    \label{fig:2_logit_shift}
    \vspace{-1.5mm}
\end{wrapfigure}

\subsection{Isolating the Objective from the Multi-Step Structure}
Evaluating the total contrast between the generation and baseline conditions ($G-D$) conflates the mere presence of a multi-step prompt structure (Step 2) with the effect of the generation objective. To isolate the cause, we utilize the Neutral 2-step condition ($N$) as a control. The determination relies on the magnitude and direction of the generation contrast ($\mathrm{ABS}^{(GD)}$, capturing the generation objective) against the structural contrast ($\mathrm{ABS}^{(ND)}$, capturing the Step 2 presence effect).

Figure~\ref{fig:2_logit_shift} demonstrates that the generation objective is responsible for the \textit{Apply bias} across all cells. The \textit{ABS} attributed to generation, $\mathrm{ABS}^{(GD)}$ is heavily positive, with its lower confidence bounds remaining strictly above zero. In contrast, the shift induced merely by structural multi-tasking, $\mathrm{ABS}^{(ND)}$, hovers near zero and frequently trends negative. This confirms that the \textit{Apply shift} is not caused by the mere presence of a multi-step prompt, but by the \textit{generation objective} itself.

\paragraph{Apply-Direction Errors Emerge at the Response Level}
We additionally analyze the behavioral impact of the generation objective by framing the model's preference application as a signal detection task \citep{green1966signal,stanislaw1999calculation}. Specifically, we track changes in the False Alarm rate (the frequency of incorrectly deciding to \textit{Apply} a preference that should be suppressed) alongside the Hit rate (the frequency of correctly deciding to \textit{Apply} an applicable preference). Observing how these rates shift reveals that Apply-direction decision errors emerge at the behavioral level across nearly all model-dataset pairs. Crucially, the simultaneous increase in both metrics points to an Apply-directed response bias rather than a general loss of discriminative ability. Detailed behavioral results, including the formal confusion matrix, are provided in Appendix \ref{app:apply_direction_errors}.

\subsection{Ruling Out Sensitivity Loss and Cascade}
\label{sec:shift_analysis}
\input{tables/Table6}
We next rule out two alternative explanations for this shift: error propagation across preference slots (\textit{a sequential cascade}) and a degradation in the model's ability to discriminate (\textit{a loss of sensitivity}).

\paragraph{Structural Isolation from Sequential Cascade} To verify whether the shift is an artifact of sequential error propagation, we fix the prefix to the ground-truth (gold) labels across all three conditions, so that no divergent histories arise before the current slot. The shift under this setup ($\mathrm{ABS}^{(\text{GD, gold})}$) is therefore driven by the generation objective alone, independent of past errors. To quantify the potential impact of cascade, we further measure \textit{Cascade Amplification} ($\text{CascadeAmp} = \mathrm{ABS}^{(\text{GD, pred})} - \mathrm{ABS}^{(\text{GD, gold})}$). As shown in Table~\ref{tab:abide_q3_results}, \textit{Cascade Amplification} varies inconsistently across models and datasets rather than exhibiting a uniform compounding pattern, indicating that sequential cascading is neither the root cause of the shift nor a reliable amplifier.

\paragraph{Preservation of Sensitivity}
We next examine whether the score shift stems from a collapse in the model's ability to distinguish between \textit{Apply} and \textit{Suppress} conditions. Across virtually all model-dataset pairs, the change in evaluation sensitivity ($\Delta \mathrm{AUC}^{(\text{GD, gold})}$) remains within the pre-defined statistical noise margin $\epsilon_{\mathrm{AUC}}$. 
This preservation of AUC confirms that the model's fundamental sensitivity remains intact under the generation objective, ruling out sensitivity loss and 
proving that the shift is driven purely by response bias.



\subsection{Ruling Out Prompt Artifacts}
\label{sec:prompt_artifacts}

To further validate that the \textit{Apply bias} stems fundamentally from the generation objective rather than prompt artifacts, we conduct ablation studies on the Ministral-3-14B-Instruct model using two variations: a \textbf{simplified generation objective ($G_2$)} and an \textbf{arithmetic neutral objective ($N_2$)}. The $G_2$ variant condenses the instruction and removes explicit ``Apply'' and ``Suppress'' terminology to rule out the possibility of lexical triggers. Conversely, the $N_2$ variant replaces the neutral fixed-sentence output with an arithmetic computation to measure the effect of multi-step task complexity. The exact prompt configurations for all variations are detailed in Appendix~\ref{appendix:ablation-prompt}.

Our results demonstrate a highly consistent pattern: regardless of the specific prompt variations, the bias induced by the generation objective remains substantially larger than that of the neutral baselines. This holds on all three datasets, with $ABS^{(G_2D)}$ exceeding $ABS^{(ND)}$ and $ABS^{(GD)}$ exceeding $ABS^{(N_2D)}$ in every case (on average, $0.89$ vs. $0.09$ and $1.51$ vs. $0.47$, respectively). Furthermore, across these variations, $\Delta AUC$ strictly remains within the noise margin (e.g., $-0.012$ to $-0.001$ under $G_2$, against margins of 0.029--0.041), indicating that the model's underlying sensitivity is robustly preserved. Detailed experimental results can be found in Appendix~\ref{sec:ablation_result}. 

\section{Reversing the Shift Repairs the Behavior}
Section~\ref{sec:abide} traced the decision failure to an Apply-directed shift in the decision score, induced by the answer-generation objective while sensitivity is preserved. If this shift drives leakage, removing it should restore the decisions and, through them, the responses. We test this in two ways, by subtracting the estimated shift from the decision score at decoding time (\S\ref{sec:debias}), and by removing the generation objective from the decision altogether with a separate decision call, which we call Des2Gen (\S\ref{sec:des2gen}).

\subsection{Debiasing the Decision Score}
\label{sec:debias}
To rigorously test whether the diagnosed bias directly causes decision failures, we introduce a \textbf{debiasing pipeline} as an analytical intervention. We first estimate a dataset- and model-specific bias scalar ($\lambda$) from the validation split used to determine the change criteria, defined as 
$$\lambda = ABS_{val}^{GD} = \frac{1}{2}(\Delta_{A_{val}}^{GD}+\Delta_{S_{val}}^{GD})$$
During the actual generation phase, before the model commits to an applicability decision, we subtract this frozen $\lambda$ from the generation condition score ($s_G$) to obtain the corrected score $\tilde{s}_G = s_G - \lambda$. The model then decodes the applicability label based on this corrected score and continues generating the response.

As shown in Table~\ref{tab:debias_results_all}, artificially reversing the score shift improves the model's selectivity. We observe a substantial increase in \textit{SR} ($\uparrow$) and overall \textit{decision accuracy} ($\uparrow$). Even though there is a loss in \textit{AR}, increasing the overall decision accuracy generally leads to improved downstream behavior. 
Crucially, this debiased decision directly improves the final generation: \textit{PLR} decreases significantly, while \textit{PFR} remains largely preserved. This result causally links the generation-induced apply bias to preference leakage, supporting that restoring the decision formation step corrects the final behavior.

\input{tables/Table8}

\subsection{Decoupling the Decision from Generation}
\label{sec:des2gen}
Alternatively, we can bypass the generation-induced apply bias entirely by decoupling the two objectives. In the \textbf{Des2Gen} approach, we instruct the model to perform the decision task in an isolated, separate call. We then inject these externally predicted applicability decisions directly into the prompt for the generation task.

By supplying the model with its own high-quality decision labels—formed without the interference of the generation objective—we observe a significant improvement in final generation performance. (Table~\ref{tab:debias_results_all}) The model successfully contextualizes the provided decisions, applying and suppressing preferences as instructed. The improvement in overall response quality indicates that the primary bottleneck driving over-personalization lies in the decision formation phase under the generation objective, rather than a fundamental inability to generate context-appropriate responses.

\section{Related Work}
\paragraph{Over-personalization} Personalized LLMs must decide which stored preferences to surface or suppress per context; failing to do so causes over-personalization. Recent benchmarks quantify this failure: OP-BENCH~\citep{hu2026op} evaluates categories like irrelevance and sychancy, BENCHPRES \citep{yoon2026benchpres} measures context-aware preference selectivity across communication norms, and RPEVAL \citep{feng2026does} assesses how irrelevant memories interfere with intent understanding. However, these benchmarks solely diagnose failures behaviorally from final responses. We complement this effort by isolating the specific processing stage where this failure occurs and identifying its underlying mechanism.
\paragraph{Over-reliance on inapplicable context} Over-personalization belongs to a broader family of failures in which models rely on context that should not shape the answer, such as irrelevant passages that distract reasoning~\citep{shi2023large} or user views that preference-tuned models tend to echo~\citep{sharma2024towards, shapira2026rlhf}. These failures are largely characterized at the output level. We instead locate the cause inside the model, showing that the answer-generation objective shifts the decision criterion toward Apply while discriminability is preserved, even without any expressed opinion or pressure from the user.

\section{Conclusion}
\label{sec:conclusion}
We studied why personalized LLMs often apply preferences the context rules out, and localized the failure by measuring knowing, decision, and generation separately. We found that the applicability signal survives into generation and wrong decisions are executed faithfully, which leaves the decision stage as the point where selectivity breaks. Using ABIDE, we showed that the generation objective shifts the decision score toward Apply while discriminability is preserved, and that subtracting a single estimated bias scalar reverses the shift and reduces leakage. Because this mechanism concerns a decision made alongside a generation objective rather than preferences as such, we expect similar shifts wherever a model must judge and generate in the same prompt.

\subsection*{AI Use Statement}
We used generative AI tools to identify literature and assist with manuscript drafting and editing.
The authors reviewed AI-assisted material, checking sources and revising the text for accuracy and clarity.
We take responsibility for the content of this work, including all AI-assisted text and claims.

\bibliography{iclr2027_conference}
\bibliographystyle{iclr2027_conference}
\newpage
\appendix
\section*{Appendix}

\section{Benchmark Integration Details}
\label{sec:appendix_benchmark}
\paragraph{Label Unification.}
To create a cohesive evaluation pipeline for contextual applicability, we standardized the labels across BenchPreS and RPEval.
\begin{itemize}[leftmargin=*, itemsep=2pt, parsep=0pt]
    \item \textbf{BenchPreS:} The native binary labels map directly to our framework. We mapped $g(t,a)=1$ to \textit{Apply} and $g(t,a)=0$ to \textit{Suppress}.
    \item \textbf{RPEval:} We mapped the original \textit{Support} (preferences that enrich the response but are not strictly required) and \textit{Dominate} (preferences that must be reflected) labels to \textit{Apply}. The \textit{Ignore} label, indicating preferences irrelevant to the current query, was mapped to \textit{Suppress}. Since our framework evaluates selectivity as a binary decision—whether a preference should be reflected in the current context rather than how strongly—both \textit{Support} and \textit{Dominate} lie on the \textit{Apply} side, and preserving the strength distinction would only confound the selectivity metrics.
\end{itemize}
\paragraph{RPEval Translation and Refinement.}
The original RPEval dataset was provided in Chinese. To support English evaluation, we translated the dataset using  GPT-5.4.
Rather than translating every instance independently, we first extracted all unique sentences to create a sentence-level translation dictionary. This approach ensured consistent phrasing across identical expressions and mitigated potential hallucinations.
Finally, manual verification was conducted to correct omissions, typos, and mistranslations.

\paragraph{Benchmark Statistics.}
The detailed statistics of the unified benchmarks, including the number of items, total preferences, and class distributions (\textit{Apply} vs. \textit{Suppress}), are summarized in Table~\ref{tab:benchmark_statistics}.

\input{tables/statistic}

\section{Implementation Details}
\label{appen:imple}

\textbf{API Access and Infrastructure}
We conducted our experiments using a mix of API and local environments. Response generation for the target models, their thinking-enabled variants, and the LLM judge were all accessed via the openRouter API. For tasks requiring local execution, we utilized NVIDIA RTX 6000 Ada Generation GPUs (48 GB). Specifically, the dedicated Ministral-3 reasoning checkpoints were served locally using vLLM (v0.28.0) with a 32,768 context length. Furthermore, to extract full-vocabulary logits for our scoring analysis, we ran the models locally with Hugging Face Transformers (v5.14.1) and PyTorch (v2.10.0). Checkpoints that exceeded a single GPU's memory were sharded using \texttt{device\_map="auto"}.

\textbf{Models}
We evaluate four open-weight instruction-tuned models: Ministral-3-8B and Ministral-3-14B (Mistral AI, Instruct-2512), Qwen3.5-27B (Alibaba), and Gemma-4-31B-it (Google). For the reasoning analysis, we additionally evaluate the dedicated reasoning checkpoint (Ministral-3-14B-Reasoning-2512) and Gemma-4-31B-it with its native thinking mode enabled. We use GPT-5.4 (openAI) both as the LLM judge for response-level metrics and as the extractor for reasoning-trace analysis.

\textbf{Decoding and Prompting}
All target-model calls use temperature 0.0 and a maximum of 2048 new tokens. Because Qwen3.5 enables thinking by default, we explicitly disable thinking for every hybrid-thinking model, so that the response-level tasks and the logit-level scoring measure the same non-thinking operating mode. The dedicated reasoning checkpoints are run with their recommended temperature of 1.0, a maximum of 16000 new tokens, and their official reasoning system prompt prepended to the task prompt. All prompts and task templates are described in Appendix~\ref{appen:prompt}.

\textbf{LLM-as-a-Judge}
Each gold preference is decomposed by GPT-5.4 into an atomic checklist of yes/no questions, each verifying a single aspect of the preference. Given the checklists of the preferences that should be applied, the judge scores the Preference Fulfillment Rate (PFR; 1--10, higher is better); given those of the preferences that should be suppressed, it scores the Preference Leakage Rate (PLR; 1--10, lower is better). All judge calls use temperature 0.0. 

\textbf{Logit-Level Apply-Bias Measurement}
For the D (direct decision), N (neutral Step 2), and G (decision + answer) conditions (Section~\ref{sec:subabide}), we read full-vocabulary raw logits at the Step 1 decision position of each preference slot, without top-$k$ log-probabilities or temperature scaling, and Step 2 is never generated. The Step 1 output format is teacher-forced and tokenized segment by segment, so the tokens that are scored are exactly the tokens fed back as context. Previous slots contain either the model's own labels (\emph{pred} prefix) or the gold label sequence shared across conditions (\emph{gold} prefix). Scoring runs in bfloat16 with batch size 1 in every condition (Ministral checkpoints are loaded from their released FP8 weights).

\textbf{Statistical Protocol}
The unit of analysis is the instance. All confidence intervals come from an instance-level cluster bootstrap with $B = 2{,}000$ replicates, in which each contrast is computed directly within every replicate. Before the main analysis, we hold out 20\% of instances as a validation split, which is used only to estimate the noise floor $\eta$ and margin defined by the noise floor and is excluded from the main estimates; the resulting margins are frozen before the main analysis. The equivalence margin is $\epsilon = 2\eta$.  As a rerun control, condition N is scored twice with a shuffled item order, and the two runs are bit-identical. We fix the random seeds for bootstrap resampling, the validation split, and the rerun order.

\section{Impact of Test-Time Scaling on Preference Leakage}
\label{sec:appendix_test_time_scaling}

In Section~\ref{sec:rq1} of the main text, we posit that over-personalization is not simply a consequence of the model underthinking its decision. To empirically validate whether extending the reasoning budget can resolve preference leakage, we conducted an additional test-time scaling experiment. We utilized the \textbf{GPT-OSS-20B} model and artificially extended its reasoning length at decoding time across three tiers: \textit{low}, \textit{medium}, and \textit{high}. We then evaluated these variants using our unified pipeline across BenchPreS, RPEval-Ex (Explicit), and RPEval-Im (Implicit).

Table \ref{tab:test_time_scaling} presents the Preference Fulfillment Rate (PFR) and Preference Leakage Rate (PLR) across the different reasoning budgets. The results clearly demonstrate that scaling up the reasoning budget at test time fails to meaningfully mitigate preference leakage. Across all three datasets, the PLR remains stubbornly high even at the \textit{high} reasoning setting (e.g., maintaining a PLR of $8.06$ on BenchPreS and $7.25$ on RPEval-Ex). Concurrently, the PFR remains largely stable regardless of the reasoning length. 

These findings reinforce our core claim: over-personalization is driven by generation-induced Apply bias rather than a mere lack of reasoning capacity. Simply forcing the model to ``think longer'' does not correct this bias.

\input{tables/test_time_scaling}

\section{Extended Probing Methodology and Results}
\label{sec:appendix_probing}

\subsection{Experimental Design: Three Evaluation Views}
To rigorously test whether the selectivity axis $w$ exists within the generation context, we evaluate our linear probes across three distinct views:
\begin{itemize}[leftmargin=*, itemsep=2pt, parsep=0pt]
    \item \textbf{Within-decision (Sanity Check):} The probe is trained and tested on the decision task activations. This verifies that the probe can successfully learn the applicability decision in an explicit context.
    \item \textbf{Within-generation:} The probe is trained and tested entirely on the generation task activations. This confirms the inherent presence of applicability information within the generation states.
    \item \textbf{Cross:} The probe is trained on the decision activations and evaluated on the generation activations. High performance in this view indicates that the exact applicability axis learned during decision transfers to generation.
\end{itemize}

\begin{figure}[h]
    \centering
    \includegraphics[width=\textwidth]{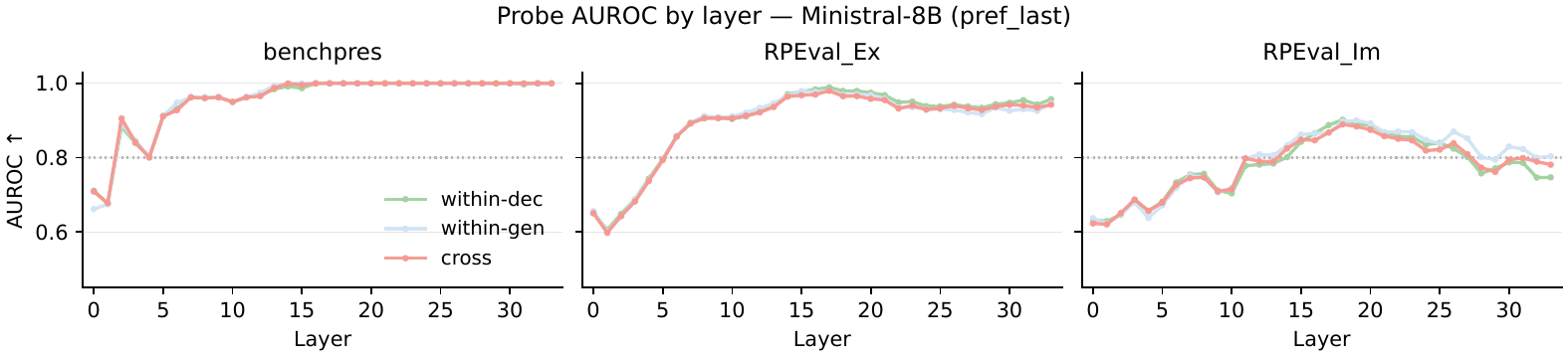} \\
    \vspace{-0.3cm} 
    
    \includegraphics[width=\textwidth]{figures/layer_auroc_Ministral-14B.pdf} \\
    \vspace{-0.3cm}
    
    \includegraphics[width=\textwidth]{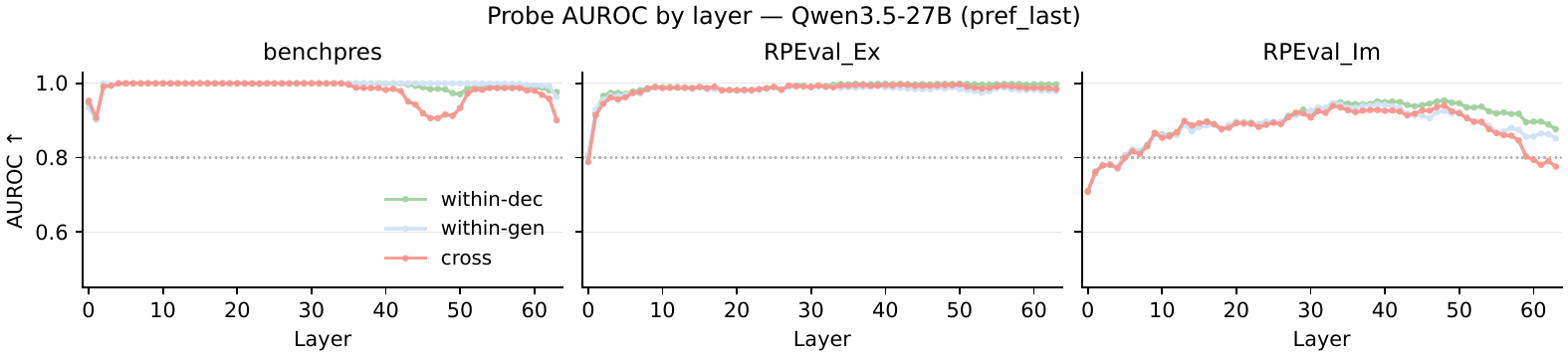} \\
    \vspace{-0.3cm}
    
    \caption{Layer-wise probe AUROC across different models. The linear applicability axis learned from the decision context robustly transfers to the generation context (\textit{cross}) in the middle-to-late layers. This trend holds consistently across all evaluated models, demonstrating that the internal knowledge of preference applicability is preserved during the generation process.}
    \label{fig:layer_auroc_all}
\end{figure}

\subsection{Results}
Across all evaluated models (including Ministral-8B, Ministral-14B, and Qwen3.5-27B), the probing results consistently demonstrate high performance. Specifically, the \textit{cross}-AUROC scores robustly exceed 0.80 in the middle-to-late layers across all datasets. This confirms that the internal applicability axis generalizes universally across different model scales and architectures.

\begin{figure}[h]
    \centering
    \includegraphics[width=\textwidth]{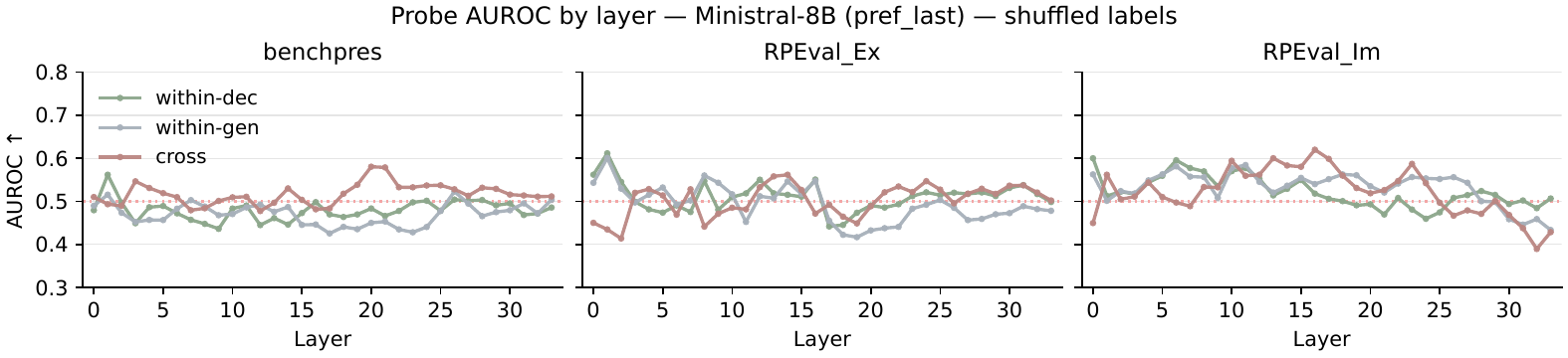} \\
    \vspace{-0.3cm} 
    
    \includegraphics[width=\textwidth]{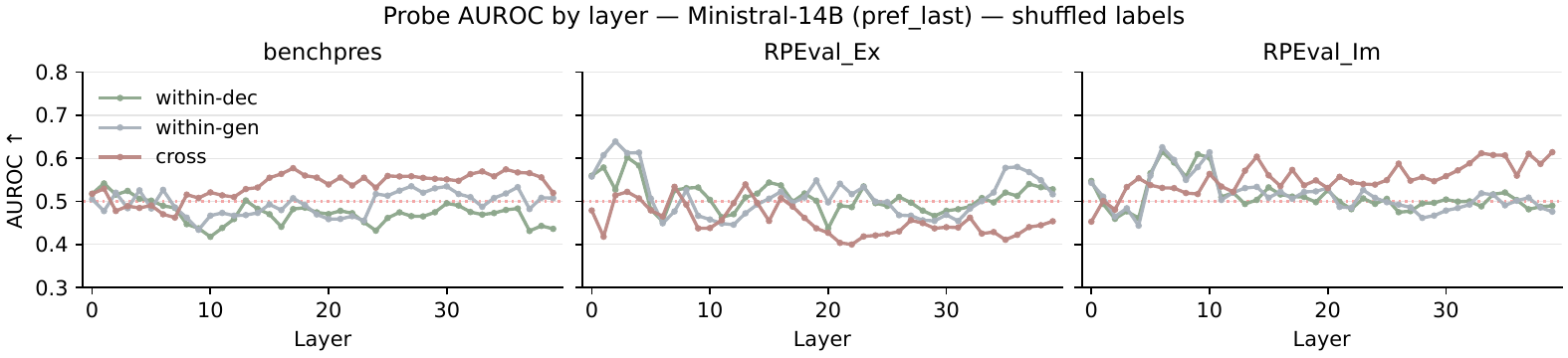} \\
    \vspace{-0.3cm}
    
    \includegraphics[width=\textwidth]{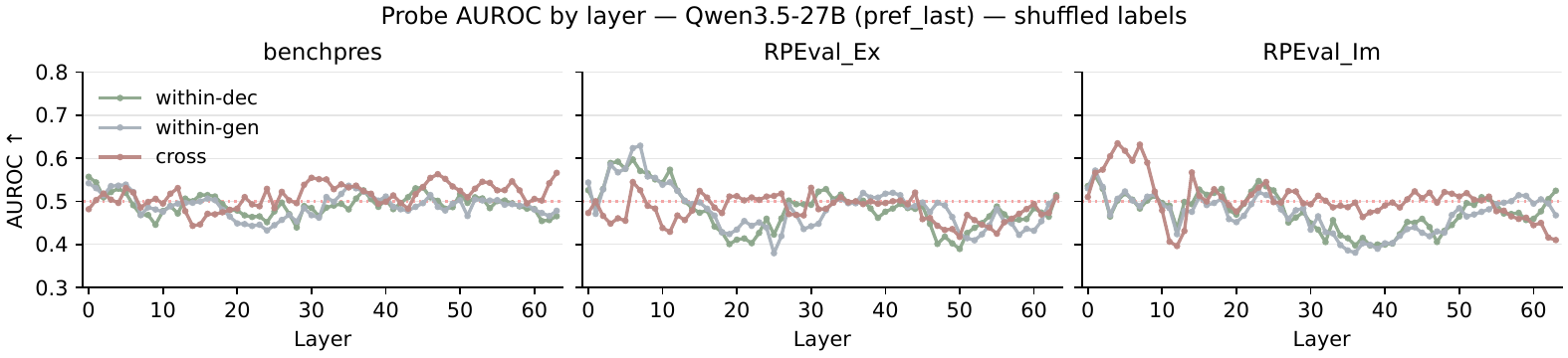} \\
    \vspace{-0.3cm}

    \caption{Layer-wise probe AUROC for the shuffled-label control across different models. When linear probes are retrained with randomized labels under our strict group holdout splits, the AUROC scores correctly drop to the random chance level of $\sim 0.5$ across all layers and evaluation views. This confirms that the high performance observed in our primary results is driven by genuine internal applicability knowledge rather than data contamination or lexical memorization.}
    \label{fig:layer_auroc_shuffled}
\end{figure}

\subsection{Shuffled-Label Control and Contamination Prevention}
To ensure our probes capture true applicability knowledge rather than simply memorizing surface-level evidence texts, we employ a strict train/test holdout strategy. We utilize a Group K-Fold Cross Validation approach where neither the exact item nor the identical preference text can appear in both the training and validation folds. 

As a further robustness check against data contamination, we evaluate a shuffled-label control \citep{hewitt-liang-2019-designing}. When the probes are retrained from scratch on randomized labels under our group holdout splits, the control AUROC correctly drops to the random chance level of $\sim 0.5$ (see Figure~\ref{fig:layer_auroc_shuffled}). This baseline confirms the absence of data contamination, demonstrating that our high standard AUROC scores (shown in Figure~\ref{fig:layer_auroc_all}) reflect genuine, generalizable internal knowledge rather than lexical artifacts.

\begin{figure*}[t]
    \centering
    
    \includegraphics[width=0.82\linewidth]{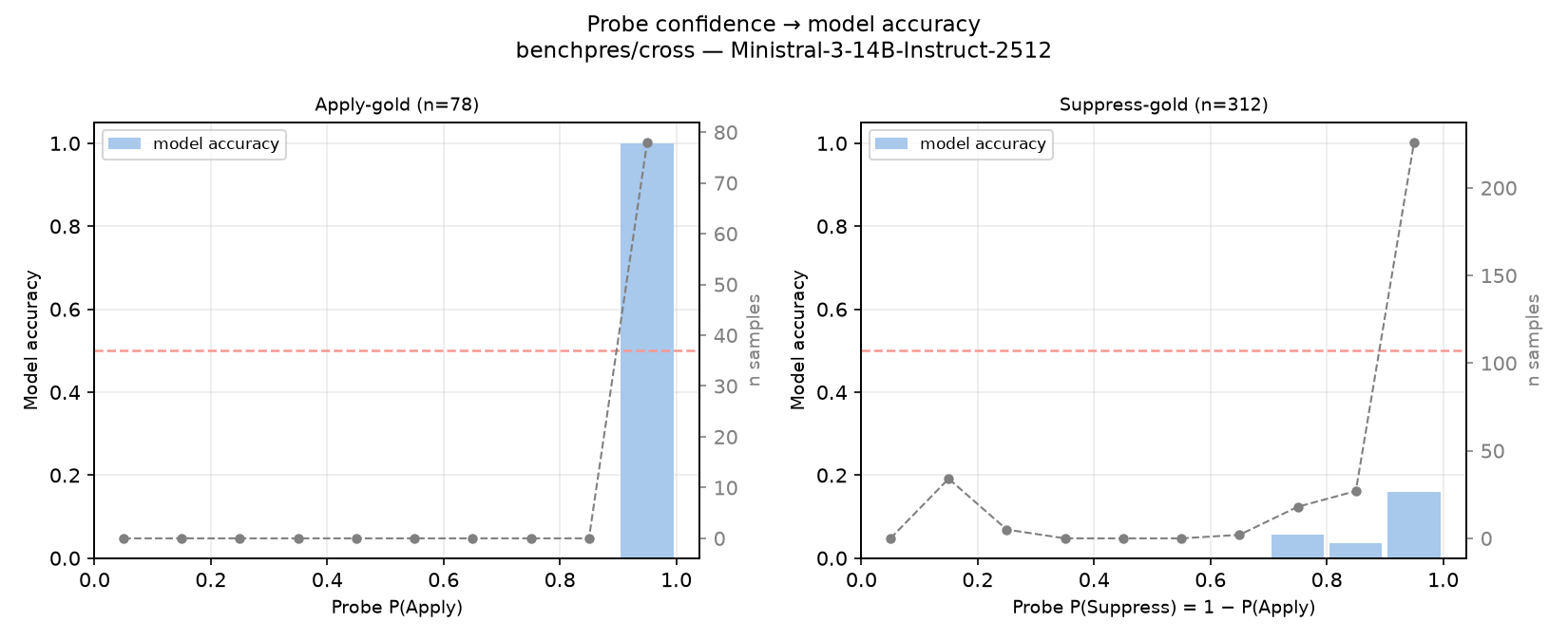}
    \vspace{-2mm}
    
    \includegraphics[width=0.82\linewidth]{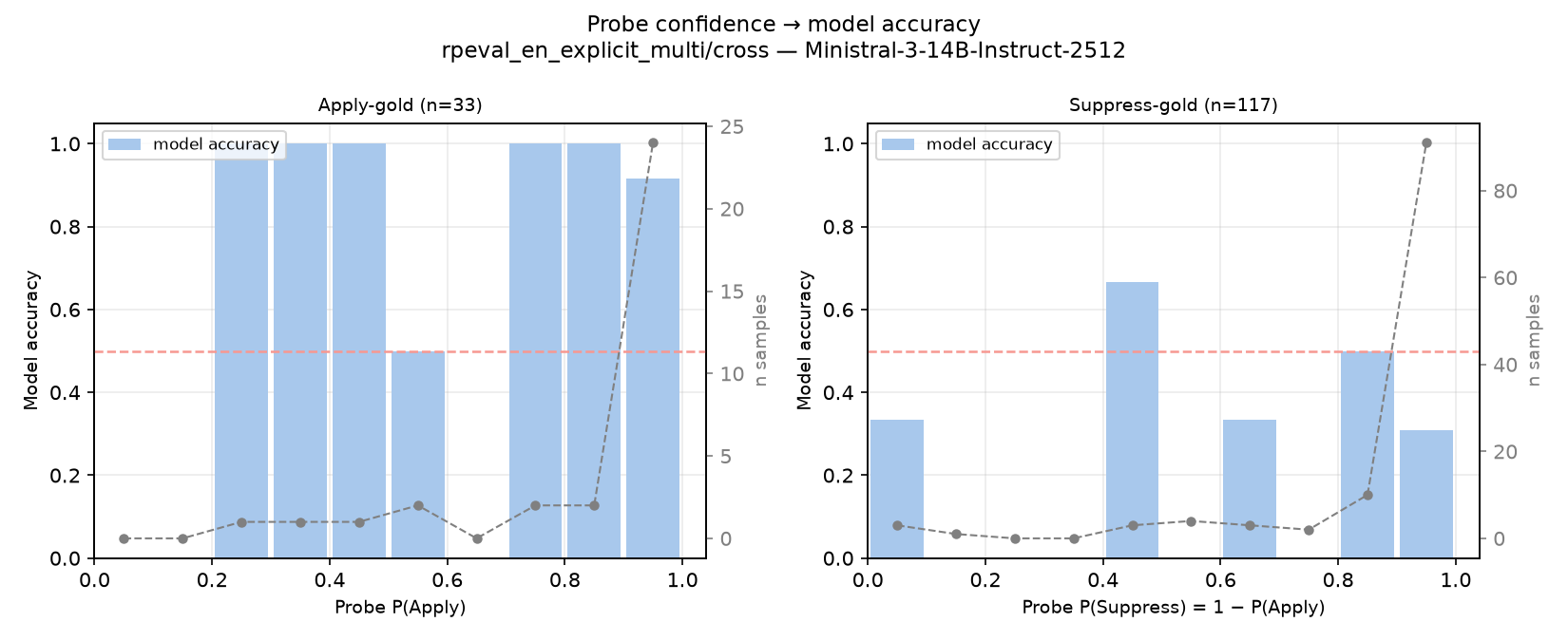}
    \vspace{-2mm}
    
    \includegraphics[width=0.82\linewidth]{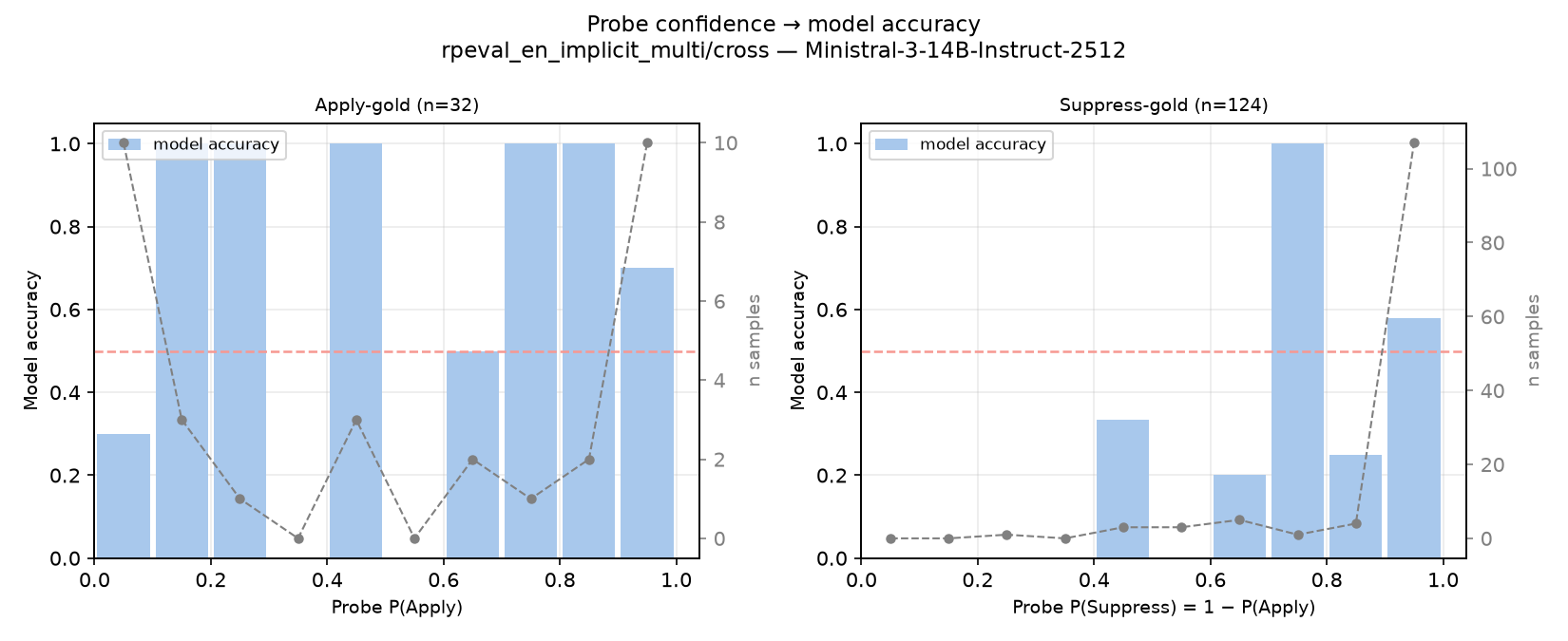}
    \vspace{-2mm}
    
    \caption{
    Probe confidence versus model behavioral accuracy for Ministral-3-14B-Instruct across \textsc{BenchPreS}, RPEval-Ex, and RPEval-Im. 
    Results are shown separately for Apply- and Suppress-gold preferences, with examples binned by probe confidence. 
    Bars indicate model accuracy within each confidence bin, while the dashed gray line indicates the number of samples in each bin.
    }
    \label{fig:probe_confidence_behavior}
    \vspace{-3mm}
\end{figure*}

\paragraph{Probe confidence vs.\ behavioral accuracy.}
As an additional diagnostic, we examine whether the confidence of the linear probe predicts the model's actual behavioral correctness. 
For each preference, we take the probe probability assigned to the gold applicability label---$P(\text{Apply})$ for Apply-gold examples and $P(\text{Suppress}) = 1-P(\text{Apply})$ for Suppress-gold examples---and bin examples according to this confidence. 
Within each bin, we compute the model's behavioral accuracy with respect to the gold Apply/Suppress label.

As shown in Figure~\ref{fig:probe_confidence_behavior}, we do not observe a consistent monotonic relationship between probe confidence and behavioral accuracy across datasets. 
In particular, for Suppress-gold examples, high-confidence probe predictions can coexist with low behavioral accuracy, indicating that strongly decodable applicability information does not necessarily translate into correct downstream behavior. 
The relationship is less informative for Apply-gold examples, where behavioral accuracy is often near ceiling and the examples are concentrated in a small number of confidence bins. 
Overall, these results further caution against interpreting probe confidence itself as a direct predictor of model behavior: the presence, or even strength, of linearly decodable applicability information does not guarantee that the model will faithfully use that information in its output.

\section{Criteria for Determining Significant Change in ABIDE}
\label{sec:appendix_margin}
To distinguish whether the observed changes in decision scores stem from simple measurement error or represent a genuine response bias, we established the following margin-based criteria. 

\paragraph{Margin Calculation and Freeze}
All margins were calculated exclusively on a separated validation split (20\% of the total dataset). The computed margins were strictly frozen prior to the main analysis to ensure the reliability of the evaluation and prevent threshold tuning. To estimate the underlying noise, we calculated the following two floors :
\begin{itemize}[leftmargin=*, itemsep=2pt, parsep=0pt]
    \item \textbf{Rerun floor ($\eta_{m}^{\mathrm{rerun}}$):} 
    We measured the inherent noise that occurs when rerunning the same prompts under identical conditions.
    \item \textbf{Paired precision floor ($\eta_{m}^{\mathrm{paired\_precision}}$):} 
    This was calculated by resampling the paired contrast at the item level within the validation split. 
    We performed 2,000 bootstrap iterations to construct a distribution of absolute differences between 
    the bootstrap mean and the overall validation mean 
    ($\lvert \text{bootstrap mean} - \text{overall validation mean} \rvert$), 
    taking the 95th percentile of this distribution as the floor value.
\end{itemize} The final base floor ($\eta_{m}^{base}$) was conservatively defined as the maximum of these two measurements.
\begin{equation}
\eta_{m}^{base} = \max(\eta_{m}^{rerun}, \eta_{m}^{paired\_precision}), \quad m \in \{ABS, AUC\}
\end{equation}
The default margin ($\epsilon$) applied to the main analysis was set to twice this base floor ($2\eta_{m}^{base}$).

\paragraph{Decision Rules}
We interpret the internal decision shifts using the following criteria:
\begin{itemize}[leftmargin=*, itemsep=2pt, parsep=0pt]
    \item $ApplyBiasShift^{(gold)} > 0$, and the lower bound of the confidence interval (CI) must be greater than 0.
    \item $\Delta AUC^{(gold)} \ge -\epsilon_{AUC}$, which implies that the model's fundamental discrimination capability (sensitivity is largely preserved.
\end{itemize}

\subsection{Apply-Direction Errors Emerge at the Response Level}
\label{app:apply_direction_errors}

We analyze the behavioral impact of the generation objective by framing the model's preference application as a signal detection task \citep{green1966signal,stanislaw1999calculation} (Table \ref{tab:apply_suppress_confusion}). In this framework, our previously defined Apply Recall (AR) is mathematically equivalent to the Hit Rate, while Suppress Recall (SR) maps directly to the Correct Rejection Rate.
\begin{table}[h]
\vspace{-2mm}
\centering
\caption{The $2 \times 2$ confusion matrix for the Apply/Suppress decision task.}
\vspace{-3mm}
\label{tab:apply_suppress_confusion}
\scriptsize
\begin{tabular}{lcc}
\toprule
& \textbf{Prediction = Apply} & \textbf{Prediction = Suppress} \\
\midrule
\textbf{Gold Class = Apply} & $H$ (Hit) & Miss \\
\textbf{Gold Class = Suppress} & $F$ (False Alarm) & Correct Rejection \\
\bottomrule

\end{tabular}
\vspace{-2mm}
\end{table}

To quantify the behavioral shift when transitioning from Direct Decision ($D$) to Decide+Answer ($G$), we measure the change in False Alarm rate, $\Delta F = F_G - F_D$, as our primary metric for Apply-direction errors, alongside the companion change in Hit rate, $\Delta H = H_G - H_D$. 

Under Signal Detection Theory (SDT)~\citep{green1966signal,stanislaw1999calculation}, tracking both metrics helps distinguish the underlying mechanism behind performance degradation: a loss of sensitivity tends to be associated with increased False Alarms and decreased Hits ($\Delta F > 0, \Delta H < 0$), whereas a response bias tends to be associated with both metrics rising together ($\Delta F > 0, \Delta H > 0$).

\begin{figure}[h]
    \centering
    \vspace{-5mm}
    \includegraphics[width=\columnwidth]{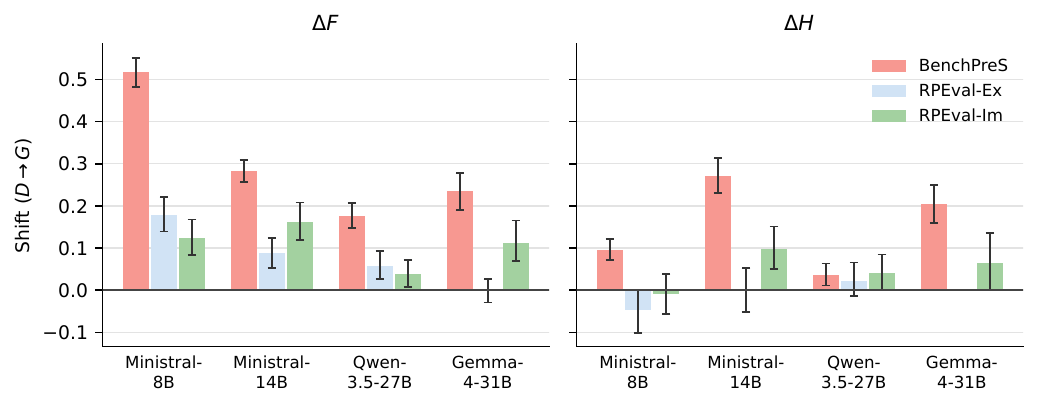}
    \vspace{-5mm}
    \caption{Behavioral shift in the Apply direction across models and datasets. Error bars indicate the 95th-percentile bootstrap margin from validation.}
    \label{fig:q1_behavioral_shift}
    \vspace{-3mm}
\end{figure}

As shown in Figure~\ref{fig:q1_behavioral_shift}, Apply-direction decision errors are supported across nearly all model-dataset pairs. $\Delta F$ and $\Delta H$ generally increase together, and at the behavioral level, this co-occurrence provides evidence consistent with an Apply-directed response bias rather than selectivity. 
The sole behavioral exception is Gemma-4-31B on RPEval-explicit, where the discrete behavior shift appears inconclusive ($\Delta F \approx 0$). Nevertheless, our score-level analysis in Section~\ref{sec:logit_shift} demonstrates that the underlying Apply-direction shift remains present even when behavioral output appears neutral.

\section{Ablation Study}

\begin{figure*}[!ht]
 \vspace{-3mm}
     \centering
     \includegraphics[width=0.495\linewidth]{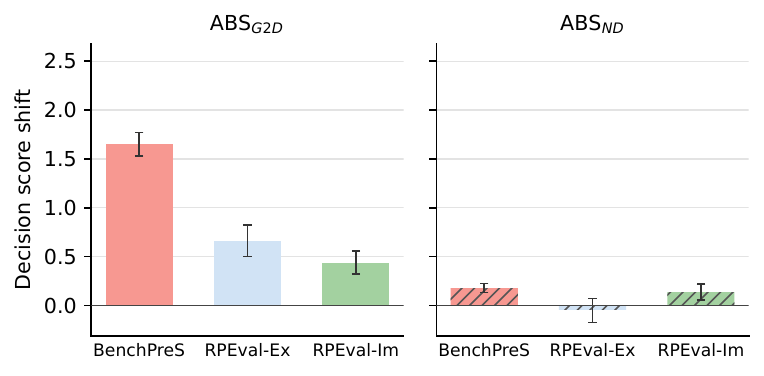}
     \hfill
     \includegraphics[width=0.495\linewidth]{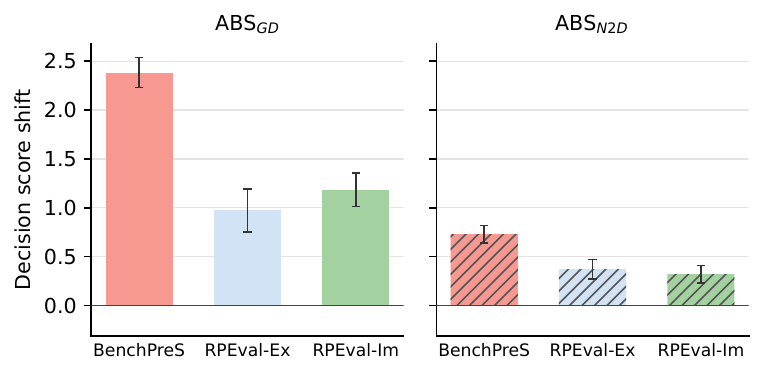}
     \vspace{-4mm}
     \caption{Robustness of generation-induced Apply bias to prompt variations. Generation objectives consistently induce larger Apply-directed decision-score shifts than neutral controls.}
     \label{fig:decision_shift}
     \vspace{-3mm}
\end{figure*}

\subsection{Detailed Ablation Results}
\label{sec:ablation_result}

To further validate that the Apply bias stems fundamentally from the generation objective rather than prompt artifacts, we evaluate two prompt variations: a simplified generation objective (G2) and an arithmetic neutral objective (N2). The G2 variant condenses the instruction and removes explicit ``Apply'' and ``Suppress'' terminology to rule out the possibility of lexical triggers, and the N2 variant replaces the fixed-sentence output of the neutral baseline (N) with an arithmetic computation to measure the effect of multi-step task complexity.

Figure~\ref{fig:decision_shift} shows the decision-score shift for each contrast. On every dataset, the generation-objective contrasts ($ABS^{(G2D)}$, $ABS^{(GD)}$) exceed their neutral counterparts ($ABS^{(ND)}$, $ABS^{(N2D)}$). Removing the explicit terminology reduces the shift relative to G but does not eliminate it, and the arithmetic step induces a modest shift that remains well below that of G.

\input{tables/ablationQ1}

\input{tables/ablationQ2}

Table~\ref{tab:q1_ablation} and Table~\ref{tab:abide_q3_ablation_results} present the detailed evaluation metrics for G2 at the behavioral and decision-score levels, respectively.

\section{Prompt Templates}
\label{appen:prompt}

\subsection{Baseline Prompt}
\label{appendix:baseline-prompt}

\begin{itemize}
    \item \textbf{Direct Decide prompt }: Table~\ref{prompt:direct-decision}
    \item \textbf{Direct Generation prompt}: Table~\ref{prompt:direct-generation}
    \item \textbf{Decide+Answer prompt}: Table~\ref{prompt:D+A}
    \item \textbf{Latent Decide+Answer}: Table~\ref{prompt:latent D+A}
    \item \textbf{Direct Decide prompt }: Table~\ref{prompt:direct-decision-implicit}
    \item \textbf{Direct Generation prompt}: Table~\ref{prompt:direct-generation-implicit}
    \item \textbf{Decide+Answer prompt}: Table~\ref{prompt:D+A-implicit}
    \item \textbf{Latent Decide+Answer prompt}: Table~\ref{prompt:latent-D+A-implicit}
\end{itemize}

\subsection{Judge Model(GPT-5.4) Prompt}
\label{appendix:judge-prompts}

\begin{itemize}
    \item \textbf{Checklist Decomposer prompt }: Table~\ref{prompt:decomposer}
    \item \textbf{Preference Fulfillment Judge (PFR) prompt}: Table~\ref{prompt:pfr-judge}
    \item \textbf{Preference Leakage Judge (PLR) prompt}: Table~\ref{prompt:plr-judge}
    \item \textbf{Reasoning Label  prompt}: Table~\ref{prompt:reasoning-label-extraction}
    \item \textbf{Generation Label Extraction prompt}: Table~\ref{prompt:generation-label-extraction}
\end{itemize}

\subsection{ABIDE Prompt}
\label{appendix:abide-prompt}
\begin{itemize}
    \item \textbf{Direct Decision prompt }: Table~\ref{prompt:Decide}
    \item \textbf{Neutral 2-step prompt}: Table~\ref{prompt:N}
\end{itemize}

\subsection{Ablation Prompt}
\label{appendix:ablation-prompt}
\begin{itemize}
    \item \textbf{Simplified Generation Objective prompt }: Table~\ref{prompt:G2}
    \item \textbf{Arithmetic Neutral Objective prompt}: Table~\ref{prompt:N2}
\end{itemize}

\input{prompts/direct_decision}
\input{prompts/direct_gen}
\input{prompts/D+A}

\input{prompts/Latent_D+A}

\input{prompts/direct_decision_implicit}
\input{prompts/direct_gen_implicit}
\input{prompts/decide+answer_implicit}

\input{prompts/latent_D+A_im}

\input{prompts/checklist_prompt}
\input{prompts/PFR}
\input{prompts/PLR}
\input{prompts/reasoning_extr}
\input{prompts/generation_Extr}

\input{prompts/decide_only}

\input{prompts/neutral}

\input{prompts/G2}

\input{prompts/N2}

\end{document}

%% file: math_commands.tex
\usepackage{amsmath,amsfonts,bm}

\def\eqref#1{equation~\ref{#1}}

\def\1{\bm{1}}

\DeclareMathAlphabet{\mathsfit}{\encodingdefault}{\sfdefault}{m}{sl}
\SetMathAlphabet{\mathsfit}{bold}{\encodingdefault}{\sfdefault}{bx}{n}



%% file: tables/Table0.tex
\begin{table}[!htbp]
\centering
\scriptsize
\renewcommand{\arraystretch}{1}
\caption{Comparison of model performance. (a) Non-reasoning models and (b) reasoning models.}
\label{tab:combined_results}

\begin{tabular*}{\textwidth}{@{\extracolsep{\fill}}llcccccccccc@{}}
\toprule

\multirow{2}{*}{\textbf{(a) Non-Reasoning models}} & \multirow{2}{*}{Dataset} 
& \multicolumn{3}{c}{Direct Decision} 
& \multicolumn{2}{c}{Direct Generation} 
& \multicolumn{3}{c}{D+A Step1} 
& \multicolumn{2}{c}{D+A Step2} \\
\cmidrule(lr){3-5}
\cmidrule(lr){6-7}
\cmidrule(lr){8-10}
\cmidrule(lr){11-12}
& & AR $\uparrow$ & SR $\uparrow$ & SS $\uparrow$ 
& PFR $\uparrow$ & PLR $\downarrow$ 
& AR $\uparrow$ & SR $\uparrow$ & SS $\uparrow$ 
& PFR $\uparrow$ & PLR $\downarrow$ \\
\midrule

\multirow{3}{*}{Ministral-3-8B-Instruct}
& BenchPreS           & 0.87 & 0.84 & 0.86 & 7.05 & 9.33 & 1.00 & 0.19 & 0.31 & 7.44 & 8.62 \\
& RPEval$_\mathrm{ex}$ & 0.80 & 0.81 & 0.80 & 6.78 & 8.05 & 0.68 & 0.58 & 0.63 & 6.44 & 5.14 \\
& RPEval$_\mathrm{im}$ & 0.74 & 0.68 & 0.71 & 5.96 & 6.93 & 0.80 & 0.50 & 0.62 & 6.09 & 5.53 \\

\cmidrule(lr){1-12}

\multirow{3}{*}{Ministral-3-14B-Instruct}
& BenchPreS           & 0.95 & 0.79 & 0.86 & 7.16 & 9.04 & 0.99 & 0.66 & 0.79 & 8.21 & 6.87 \\
& RPEval$_\mathrm{ex}$ & 0.95 & 0.75 & 0.84 & 7.03 & 8.21 & 0.86 & 0.68 & 0.76 & 7.23 & 5.04 \\
& RPEval$_\mathrm{im}$ & 0.78 & 0.66 & 0.71 & 6.36 & 7.31 & 0.77 & 0.66 & 0.71 & 6.09 & 5.21 \\

\cmidrule(lr){1-12}

\multirow{3}{*}{Qwen3.5-27B}
& BenchPreS           & 0.96 & 0.91 & 0.93 & 7.85 & 9.51 & 0.92 & 0.77 & 0.84 & 8.01 & 3.12 \\
& RPEval$_\mathrm{ex}$ & 0.98 & 0.80 & 0.88 & 7.90 & 8.87 & 0.98 & 0.73 & 0.84 & 7.88 & 5.59 \\
& RPEval$_\mathrm{im}$ & 0.82 & 0.75 & 0.78 & 7.19 & 6.21 & 0.90 & 0.65 & 0.75 & 7.09 & 5.73 \\

\cmidrule(lr){1-12}

\multirow{3}{*}{Gemma-4-31B-it}
& BenchPreS           & 0.88 & 0.99 & 0.93 & 8.23 & 9.44 & 0.64 & 0.90 & 0.75 & 6.81 & 1.83 \\
& RPEval$_\mathrm{ex}$ & 0.94 & 0.74 & 0.83 & 8.18 & 8.30 & 0.99 & 0.75 & 0.85 & 8.19 & 4.89 \\
& RPEval$_\mathrm{im}$ & 0.83 & 0.74 & 0.78 & 6.99 & 6.77 & 0.85 & 0.63 & 0.72 & 7.30 & 5.47 \\

\bottomrule
\end{tabular*}

\begin{tabular*}{\textwidth}{@{\extracolsep{\fill}}llccccccc@{}}
\toprule

\multirow{2}{*}{\textbf{(b) Reasoning models}} & \multirow{2}{*}{Dataset} 
& \multicolumn{3}{c}{Direct Decision} 
& \multicolumn{2}{c}{Direct Generation} 
& \multicolumn{2}{c}{Latent D+A (Gen)} \\
\cmidrule(lr){3-5}
\cmidrule(lr){6-7}
\cmidrule(lr){8-9}
& & AR $\uparrow$ & SR $\uparrow$ & SS $\uparrow$ 
& PFR $\uparrow$ & PLR $\downarrow$ 
& PFR $\uparrow$ & PLR $\downarrow$ \\
\midrule

\multirow{3}{*}{Ministral-3-14B-Reasoning}
& BenchPreS           & 0.84 & 0.97 & 0.90 & 7.17 & 8.70 & 7.59 & 4.96 \\
& RPEval$_\mathrm{ex}$ & 0.90 & 0.75 & 0.82 & 7.13 & 7.87 & 7.16 & 5.27 \\
& RPEval$_\mathrm{im}$ & 0.79 & 0.78 & 0.78 & 6.10 & 6.03 & 6.59 & 5.20 \\

\cmidrule(lr){1-9}

\multirow{3}{*}{Gemma-4-31B-it (think)}
& BenchPreS           & 0.96 & 0.97 & 0.96 & 8.11 & 9.57 & 7.67 & 5.08 \\
& RPEval$_\mathrm{ex}$ & 0.96 & 0.72 & 0.82 & 8.33 & 8.31 & 7.10 & 5.07 \\
& RPEval$_\mathrm{im}$ & 0.94 & 0.56 & 0.70 & 7.32 & 7.07 & 7.41 & 6.54 \\

\bottomrule
\end{tabular*}

\end{table}

%% file: tables/Table6.tex
\begin{table}[h]
\centering
\scriptsize
\caption{ABIDE analysis under the gold prefix, comparing the local decision score shift (G vs.\ D) and the Step~2 presence control (N vs.\ D) against cascade amplification across models and datasets.}
\label{tab:abide_q3_results}
\begin{tabular}{llcccc}
\toprule
\textbf{Model} & \textbf{Dataset} & $\mathbf{ABS^{(\text{GD, gold})}}$ & $\mathbf{ABS^{(\text{ND, gold})}}$ & $\mathbf{\Delta \mathrm{AUC}^{(\text{GD, gold})} / \epsilon_{\mathrm{AUC}}}$ & \textbf{CascadeAmp} \\
\midrule
\multirow{3}{*}{Ministral-3-8B} 
& BenchPreS & $+3.028$ & $+0.105$ & $-0.048$ / 0.086 & $+0.183$ \\
& RPEval-Ex & $+0.973$ & $-0.172$ & $-0.068$ / 0.122 & $-0.056$ \\
& RPEval-Im & $+1.234$ & $-0.077$ & $-0.054$ / 0.085 & $-0.274$ \\
\midrule
\multirow{3}{*}{Ministral-3-14B} 
& BenchPreS & $+2.441$ & $+0.192$ & $+0.022$ / 0.045 & $-0.058$ \\
& RPEval-Ex & $+1.041$ & $-0.082$ & $-0.025$ / 0.034 & $-0.059$ \\
& RPEval-Im & $+1.449$ & $+0.045$ & $-0.003$ / 0.061 & $-0.266$ \\
\midrule
\multirow{3}{*}{Qwen3.5-27B} 
& BenchPreS & $+1.259$ & $+0.095$ & $-0.018$ / 0.017 & $+0.285$ \\
& RPEval-Ex & $+0.689$ & $-0.049$ & $-0.011$ / 0.032 & $+0.082$ \\
& RPEval-Im & $+0.538$ & $-0.018$ & $+0.003$ / 0.034 & $+0.069$ \\
\midrule
\multirow{3}{*}{Gemma-4-31B-it} 
& BenchPreS & $+5.165$ & $-0.465$ & $-0.037$ / 0.038 & $+1.952$ \\
& RPEval-Ex & $+0.300$ & $-1.062$ & $+0.000$ / 0.015 & $-0.071$ \\
& RPEval-Im & $+1.125$ & $-1.071$ & $-0.027$ / 0.049 & $+1.256$ \\
\bottomrule
\end{tabular}
\end{table}

%% file: tables/Table8.tex
\begin{table}[h]
\centering
\scriptsize
\setlength{\tabcolsep}{3pt}
\renewcommand{\arraystretch}{0.85}

\caption{Impact of the Debias Pipeline on Decision + Answer (D+A) Performance Across Various Models. The metrics present Original $\rightarrow$ Debiased values, while Des2Gen reports generation performance after explicitly separating decision and generation.}
\label{tab:debias_results_all}

\begin{tabular}{llccc cc|cc}
\toprule
\multirow{2}{*}{\textbf{Model}}
& \multirow{2}{*}{\textbf{Dataset}}
& \multicolumn{3}{c}{\textbf{Debiased D+A (Step 1)}}
& \multicolumn{2}{c@{\hspace{12pt}}}{\textbf{Debiased D+A (Step 2)}}
& \multicolumn{2}{c}{\textbf{Des2Gen}} \\
\cmidrule(lr){3-5}
\cmidrule(lr){6-7}
\cmidrule(lr){8-9}
&
& \textbf{AR}
& \textbf{SR}
& \textbf{Acc}
& \textbf{PFR $\uparrow$}
& \textbf{PLR $\downarrow$}
& \textbf{PFR $\uparrow$}
& \textbf{PLR $\downarrow$} \\
\midrule

\multirow{3}{*}{\textbf{Ministral-3-8B}}
& BenchPreS
& $1.00 \rightarrow 0.73$
& $0.15 \rightarrow 0.71$
& $0.44 \rightarrow 0.72$
& $7.50 \rightarrow 8.12$
& $8.55 \rightarrow 6.65$
& 8.31
& 2.71 \\

& RPEval$_\mathrm{ex}$
& $0.66 \rightarrow 0.63$
& $0.55 \rightarrow 0.70$
& $0.57 \rightarrow 0.68$
& $6.51 \rightarrow 6.81$
& $5.37 \rightarrow 4.94$
& 6.98
& 4.47 \\

& RPEval$_\mathrm{im}$
& $0.80 \rightarrow 0.72$
& $0.46 \rightarrow 0.61$
& $0.53 \rightarrow 0.63$
& $5.83 \rightarrow 5.90$
& $5.84 \rightarrow 5.40$
& 5.93
& 5.98 \\

\midrule

\multirow{3}{*}{\textbf{Ministral-3-14B}}
& BenchPreS
& $1.00 \rightarrow 0.77$
& $0.62 \rightarrow 0.85$
& $0.75 \rightarrow 0.82$
& $8.25 \rightarrow 8.36$
& $7.16 \rightarrow 5.49$
& 8.39
& 3.32 \\

& RPEval$_\mathrm{ex}$
& $0.89 \rightarrow 0.84$
& $0.66 \rightarrow 0.76$
& $0.71 \rightarrow 0.78$
& $6.99 \rightarrow 7.26$
& $5.05 \rightarrow 4.72$
& 7.10
& 4.97 \\

& RPEval$_\mathrm{im}$
& $0.80 \rightarrow 0.72$
& $0.66 \rightarrow 0.77$
& $0.69 \rightarrow 0.76$
& $5.99 \rightarrow 6.06$
& $5.16 \rightarrow 4.68$
& 6.42
& 6.22 \\

\midrule

\multirow{3}{*}{\textbf{Qwen3.5-27B}}
& BenchPreS
& $0.96 \rightarrow 0.77$
& $0.80 \rightarrow 0.92$
& $0.85 \rightarrow 0.87$
& $8.59 \rightarrow 7.84$
& $3.00 \rightarrow 1.75$
& 8.75
& 1.84 \\

& RPEval$_\mathrm{ex}$
& $0.98 \rightarrow 0.95$
& $0.73 \rightarrow 0.79$
& $0.78 \rightarrow 0.83$
& $8.00 \rightarrow 8.00$
& $5.37 \rightarrow 5.03$
& 8.37
& 4.83 \\

& RPEval$_\mathrm{im}$
& $0.88 \rightarrow 0.85$
& $0.69 \rightarrow 0.73$
& $0.72 \rightarrow 0.75$
& $7.19 \rightarrow 7.14$
& $5.49 \rightarrow 5.34$
& 7.42
& 5.33 \\

\midrule

\multirow{3}{*}{\textbf{Gemma-4-31B-it}}
& BenchPreS
& $0.65 \rightarrow 0.48$
& $0.86 \rightarrow 0.95$
& $0.79 \rightarrow 0.79$
& $6.66 \rightarrow 5.93$
& $2.18 \rightarrow 1.49$
& 8.22
& 1.06 \\

& RPEval$_\mathrm{ex}$
& $0.99 \rightarrow 0.99$
& $0.75 \rightarrow 0.75$
& $0.80 \rightarrow 0.80$
& $8.16 \rightarrow 8.16$
& $4.68 \rightarrow 4.68$
& 8.10
& 5.23 \\

& RPEval$_\mathrm{im}$
& $0.85 \rightarrow 0.84$
& $0.60 \rightarrow 0.64$
& $0.65 \rightarrow 0.68$
& $7.28 \rightarrow 7.41$
& $5.53 \rightarrow 5.43$
& 7.46
& 4.93 \\

\bottomrule
\end{tabular}
\vspace{-4mm}
\end{table}

%% file: tables/statistic.tex
\begin{table}[h]

\centering

\small

\begin{tabular}{lcccc}

\toprule

\textbf{Benchmark} & \textbf{Item} & \textbf{Pref} & \textbf{Apply} & \textbf{Suppress} \\

\midrule

BenchPreS & 390 & 1,950 & 663 & 1,287 \\

RPEval-Ex & 150 & 803 & 162 & 641 \\

RPEval-Im & 150 & 804 & 163 & 641 \\

\midrule

\textbf{Total} & 690 & 3,557 & 988 & 2,569 \\

\bottomrule

\end{tabular}

\caption{Statistics of the preference benchmarks.}

\label{tab:benchmark_statistics}

\end{table}

%% file: tables/test_time_scaling.tex
\begin{table}[h]
\centering
\caption{Effect of test-time reasoning scaling on Preference Fulfillment Rate (PFR $\uparrow$) and Preference Leakage Rate (PLR $\downarrow$) using GPT-OSS-20B. Extending the reasoning budget at inference time does not alleviate preference leakage, confirming that over-personalization cannot be resolved by longer reasoning alone.}
\label{tab:test_time_scaling}
\vspace{0.2cm}
\begin{tabular}{lcccccc}
\toprule
\textbf{Model /} & \multicolumn{2}{c}{\textbf{BenchPreS}} & \multicolumn{2}{c}{\textbf{RPEval-Ex}} & \multicolumn{2}{c}{\textbf{RPEval-Im}} \\
\cmidrule(lr){2-3} \cmidrule(lr){4-5} \cmidrule(lr){6-7}
\textbf{Reasoning Budget} & PFR $\uparrow$ & PLR $\downarrow$ & PFR $\uparrow$ & PLR $\downarrow$ & PFR $\uparrow$ & PLR $\downarrow$ \\
\midrule
GPT-OSS-20B-low    & 7.51 & 8.04 & 7.10 & 7.11 & 5.37 & 3.29 \\
GPT-OSS-20B-medium & 7.52 & 8.16 & 7.22 & 7.14 & 5.73 & 3.29 \\
GPT-OSS-20B-high   & 7.49 & 8.06 & 7.26 & 7.25 & 5.53 & 3.48 \\
\bottomrule
\end{tabular}
\end{table}

%% file: tables/ablationQ1.tex
\begin{table}[h]
\centering
\caption{Q1 ablation analysis comparing apply-direction behavioral errors at the behavior level between the simplified generation objective (G2) and the decision baseline (D) of Ministral-3-14B-Instruct.}
\label{tab:q1_ablation}
\small
\begin{tabular}{lccc}
\toprule
 \textbf{Dataset} & $\boldsymbol{F_D \to F_G}$ & $\boldsymbol{\Delta F}$ \textbf{[95\% CI]} & $\boldsymbol{\Delta H}$ \textbf{[95\% CI]} \\
\midrule
BenchPreS & .208 $\rightarrow$ .454 & +.245 [.220, .271] & +.263 [.224, .306] \\
RPEval Explicit & .266 $\rightarrow$ .343 & +.077 [.045, .110] & +.023 [-.032, +.083] \\
RPEval Implicit & .307 $\rightarrow$ .367 & +.060 [.025, .096] & +.049 [.008, .095] \\
\bottomrule
\end{tabular}
\end{table}

%% file: tables/ablationQ2.tex
\begin{table}[h]
\centering
\small
\caption{ABIDE analysis comparing local decision score shift against cascade amplification across ablation prompt variant of Ministral-3-14B-Instruct. Margins ($\epsilon_{\mathrm{AUC}}$) are frozen on the validation split}
\label{tab:abide_q3_ablation_results}
\begin{tabular}{lccc}
\toprule
\textbf{Dataset} & $\mathbf{ABS^{(\text{gold})}}$ & $\mathbf{\Delta \mathrm{AUC}^{(\text{gold})} / \epsilon_{\mathrm{AUC}}}$ & \textbf{CascadeAmp} \\
\midrule
BenchPreS  & $+1.618$ & $-0.006$ / 0.034 & $-0.146$ \\
RPEval-Ex  & $+0.698$ & $-0.012$ / 0.029 & $+0.003$ \\
RPEval-Im  & $+0.350$ & $-0.001$ / 0.041 & $-0.049$ \\
\bottomrule
\end{tabular}
\end{table}

%% file: prompts/direct_decision.tex
\begin{table*}[h]
\centering
\caption{Direct Decision prompt template for classifying the contextual applicability of user preferences.}
\scalebox{0.9}{
\begin{tabularx}{\textwidth}{X}
\toprule
\textbf{Direct Decision} \\
\midrule
You are an expert at judging whether user preferences should be applied or suppressed in a given context. \\
For each preference, output exactly "Apply" or "Suppress".\\
\\
Rules:\\
"Apply" means the preference is relevant and appropriate for the current context.\\
"Suppress" means the preference is irrelevant, inappropriate, or would conflict with the context's norms or the user's actual intent.\\
\\
Output one decision for each preference in order.\\
\bottomrule
\end{tabularx}
}
\label{prompt:direct-decision}
\end{table*}

%% file: prompts/direct_gen.tex
\begin{table*}[h]
\centering
\caption{Direct Generation prompt template for generating a response using stored user preferences.}
\scalebox{0.9}{
\begin{tabularx}{\textwidth}{X}
\toprule
\textbf{Direct Generation} \\
\midrule
You will be provided with a request from a user along with stored information about the user's preferences. You should provide a helpful response to the request.\\
\\
To help you understand the user's needs and expectations, use the stored preference information to craft a response that naturally reflects what is appropriate for the user in the current context. Consider which stored information is relevant to the current situation.\\
\bottomrule
\end{tabularx}
}
\label{prompt:direct-generation}
\end{table*}

%% file: prompts/D+A.tex
\begin{table*}[h]
\centering
\caption{Decide+Answer prompt template for conditionally applying or suppressing user preferences based on context.}
\scalebox{0.9}{ 
\begin{tabularx}{\textwidth}{X}
\toprule
\textbf{Decide+Answer} \\
\midrule
\textbf{Step 1:} Decide whether user preferences should be applied or suppressed in a given context.\\
For each preference, output exactly "Apply" or "Suppress".\\
Rules:\\
"Apply" means the preference is relevant and appropriate for the current context.\\
"Suppress" means the preference is irrelevant, inappropriate, or would conflict with the context's norms or the user's actual intent.\\
Present your analysis under the header:\\
Step 1 output:\\
\{step1\_format\}\\
\\
\textbf{Step 2:} Provide a helpful response to the user's request.\\
Use the preferences you marked as relevant, and ignore those you marked as not relevant, to craft a response that naturally reflects what is appropriate for this user. \\
Your Step 2 output must contain only the content addressing the user's request — do not include your reasoning or preference classifications. \\ 
Present your response under the header:\\
Step 2 output:\\
\bottomrule
\end{tabularx}
}
\label{prompt:D+A}
\end{table*}

%% file: prompts/latent_D+A.tex
\begin{table*}[h]
\centering
\caption{Latent Decide+Answer prompt template for conditionally applying or suppressing user preferences based on context.}
\scalebox{0.9}{ 
\begin{tabularx}{\textwidth}{X}
\toprule
\textbf{Latent Decide+Answer} \\
\midrule
\textbf{Step 1:} Decide whether user preferences should be applied or suppressed in a given context.\\
For each preference, output exactly "Apply" or "Suppress".\\
Rules:\\
"Apply" means the preference is relevant and appropriate for the current context.\\
"Suppress" means the preference is irrelevant, inappropriate, or would conflict with the context's norms or the user's actual intent.\\
\textbf{Step 2:} Provide a helpful response to the user's request.\\
Use the preferences you marked as relevant, and ignore those you marked as not relevant, to craft a response that naturally reflects what is appropriate for this user. \\
Your Step 2 output must contain only the content addressing the user's request — do not include your reasoning or preference classifications. \\ 
\bottomrule
\end{tabularx}
}
\label{prompt:latent D+A}
\end{table*}

%% file: prompts/direct_decision_implicit.tex
\begin{table*}[h]
\centering
\caption{Direct Decision prompt template for implicit preferences in RPEval-Im.}
\scalebox{0.9}{
\begin{tabularx}{\textwidth}{X}
\toprule
\textbf{Direct Decision (Implicit)} \\
\midrule
You are an expert at judging whether prior conversations reveal implicit user preferences that are relevant to the current request.\\
\\
For each conversation, decide whether the implicit preferences it reveals should be applied or suppressed when generating a response to the current request.\\
For each conversation, output exactly "Apply" or "Suppress".\\
\\
Rules:\\
"Apply" means the conversation reveals implicit preferences that are relevant and appropriate for the current context.\\
"Suppress" means the conversation's implicit preferences are irrelevant, inappropriate, or would conflict with the context's norms or the user's actual intent.\\
\bottomrule
\end{tabularx}
}
\label{prompt:direct-decision-implicit}
\end{table*}

%% file: prompts/direct_gen_implicit.tex
\begin{table*}[h]
\centering
\caption{Direct Generation prompt template for implicit preferences in RPEval-Im.}
\scalebox{0.9}{
\begin{tabularx}{\textwidth}{X}
\toprule
\textbf{Direct Generation (Implicit)} \\
\midrule
You will be provided with a request from a user. You should provide a helpful response to the request.\\
\\
To help you understand the user's needs and expectations, you will be provided with prior conversation history between the user and an AI assistant. You must infer what expectations the user held in similar previous interactions from this history to craft a response that meets the user's expectations.\\
\bottomrule
\end{tabularx}
}
\label{prompt:direct-generation-implicit}
\end{table*}

%% file: prompts/decide+answer_implicit.tex
\begin{table*}[h]
\centering
\caption{Decide+Answer prompt template for implicit preferences in RPEval-Im.}
\scalebox{0.9}{
\begin{tabularx}{\textwidth}{X}
\toprule
\textbf{Decide+Answer (Implicit)} \\
\midrule
\textbf{Step 1:} Decide whether prior conversations reveal implicit user preferences that are relevant to the current request.\\
\\
For each conversation, decide whether the implicit preferences it reveals should be applied or suppressed when generating a response to the current request.\\
For each conversation, output exactly "Apply" or "Suppress".\\
\\
Rules:\\
"Apply" means the conversation reveals implicit preferences that are relevant and appropriate for the current context.\\
"Suppress" means the conversation's implicit preferences are irrelevant, inappropriate, or would conflict with the context's norms or the user's actual intent.\\
\\
Present your analysis under the header:\\
Step 1 output:\\
\{step1\_format\}\\
\\
\textbf{Step 2:} Provide a helpful response to the user's request.\\
Use the conversations you marked as relevant, and ignore those you marked as not relevant, to craft a response that naturally reflects what is appropriate for this user.\\
Your Step 2 output must contain only the content addressing the user's request --- do not include your reasoning or conversation classifications.\\
Present your response under the header:\\
Step 2 output:\\
\bottomrule
\end{tabularx}
}
\label{prompt:D+A-implicit}
\end{table*}

%% file: prompts/latent_D+A_im.tex
\begin{table*}[h]
\centering
\caption{Latent Decide+Answer prompt template for implicit preferences in RPEval-Im.}
\scalebox{0.9}{
\begin{tabularx}{\textwidth}{X}
\toprule
\textbf{Latent Decide+Answer (Implicit)} \\
\midrule
\textbf{Step 1:} Decide whether prior conversations reveal implicit user preferences that are relevant to the current request.\\
\\
For each conversation, decide whether the implicit preferences it reveals should be applied or suppressed when generating a response to the current request.\\
For each conversation, output exactly "Apply" or "Suppress".\\
\\
Rules:\\
"Apply" means the conversation reveals implicit preferences that are relevant and appropriate for the current context.\\
"Suppress" means the conversation's implicit preferences are irrelevant, inappropriate, or would conflict with the context's norms or the user's actual intent.\\
\\
\textbf{Step 2:} Provide a helpful response to the user's request.\\
Use the conversations you marked as relevant, and ignore those you marked as not relevant, to craft a response that naturally reflects what is appropriate for this user.\\
Your Step 2 output must contain only the content addressing the user's request --- do not include your reasoning or conversation classifications.\\
\bottomrule
\end{tabularx}
}
\label{prompt:latent-D+A-implicit}
\end{table*}

%% file: prompts/checklist_prompt.tex
\begin{table*}[h]
\centering
\caption{Checklist decomposition prompt used to decompose a user preference into atomic yes/no evaluation criteria.}
\scalebox{0.80}{
\begin{tabularx}{\textwidth}{X}
\toprule
\textbf{Checklist Decomposer} \\
\midrule

\textbf{\#\# Your Objective}\\
Your task is to help judge how well an AI Assistant's response satisfies a given preference by creating an evaluation checklist from the preference. Here, a \textbf{preference} refers to a requirement, guideline, or principle that a user considers when assessing the quality of an AI Assistant's response.\\
\\
\textbf{\#\# Task Details}\\
Your task is to come up with an evaluation checklist for a given preference. This checklist should be a list of questions that ask whether or not specific aspects contained within a preference were met by an AI assistant's response.\\
\\
Checklist questions should:\\
- \textbf{Be answerable by `yes' or `no'}, with `yes' meaning the response successfully met the corresponding requirement.\\
- \textbf{Be comprehensive, but concise}: all aspects directly relevant to the preference should be represented, but only clearly relevant questions should be included.\\
- \textbf{Be precise}: avoid vague wording and evaluate specific aspects directly, using the phrasing of the preference where appropriate. Avoid introducing new content not included in the preference.\\
\\
Analyse the preference before providing the checklist. Break it down into \textbf{as many atomic facts as possible}, with one question per atomic fact.\\
\\
\textbf{\#\# Response Format}\\
\\
\textbf{Analysis}\\
\\
<Explain your analysis of the preference here.>\\
\\
\textbf{Checklist}\\
\\
\{\\
\hspace{3mm}"checklist": [\\
\hspace{6mm}"<question 1>",\\
\hspace{6mm}"<question 2>"\\
\hspace{3mm}]\\
\}\\
\\
\textbf{\#\# Examples}\\
\\
Preference: "Responsibilities should be communicated with a stern and direct tone."\\
Checklist: ["Are the responsibilities communicated in a stern tone?", "Are the responsibilities communicated in a direct tone?"]\\
\\
Preference: "Training manuals should provide clear, step-by-step instructions for all procedures."\\
Checklist: ["Are the instructions clear?", "Do the instructions follow a step-by-step format?", "Does the response provide instructions for all procedures?"]\\
\\
\textbf{User message:}\\
\textbf{Preference}\\
\\
\{preference\}\\

\bottomrule
\end{tabularx}
}
\label{prompt:decomposer}
\end{table*}

%% file: prompts/PFR.tex

\begin{table*}[h]
\centering
\caption{LLM-as-a-judge prompt used to evaluate preference fulfillment for Apply preferences.}
\scalebox{0.77}{
\begin{tabularx}{\textwidth}{X}
\toprule
\textbf{Preference Fulfillment Judge (PFR)} \\
\midrule

\textbf{\#\# Your Objective}\\
\\
You are a critical and meticulous evaluator. You will be presented with a user's request to an AI assistant and the AI's response to the user. Your task is to evaluate whether the AI assistant's response satisfied the user's \textbf{personal preferences}. To help you evaluate the responses on preference, you will also be provided with an \textbf{evaluation checklist} that decomposes the preference into specific questions.\\
\\
\textbf{\#\#\# Evaluation Preferences and Checklists}\\
\\
\textbf{Preference 1}: "\{preference\_1\}"\\
\textbf{Evaluation Checklist}:\\
\hspace{3mm}- \{checklist\_item\_1\}\\
\hspace{3mm}- \{checklist\_item\_2\}\\
\\
$\cdots$\\
\\
\textbf{\#\# Instructions}\\
\\
You should write down your analysis and assessment on how well the AI assistant's response satisfies each item in your checklist. You should follow these considerations:\\
\\
- Walk through each checklist item and summarize the response's "strengths" and "weaknesses" regarding that checklist item.\\
- For each checklist item, you should consider whether the checklist item was satisfied or dissatisfied.\\
- Avoid considering aspects that are not included in the checklist. Focus only on the evaluation checklist. Ensure that your persona profile does not influence your evaluation.\\
- You should then return a score in the range of 1--10, where 1 means the response completely fails to satisfy the preferences and 10 means the response is completely perfect.\\
\\
Ensure that you follow the format given below. Avoid adding additional content that is not included in the format below.\\
\\
---\\
\\
\textbf{\#\# Output Format}\\
\\
\textbf{\#\#\# Evaluation of AI Assistant's Response}\\
\\
\textbf{Preference 1: \{preference\_1\}}\\
1. \textbf{\{checklist\_item\_1\}}: <Detailed analysis of the response's strengths and weaknesses regarding this checklist item>\\
2. \textbf{\{checklist\_item\_2\}}: <Detailed analysis of the response's strengths and weaknesses regarding this checklist item>\\
\\
$\cdots$\\
\\
\textbf{\#\#\# Evaluation Score}\\
\\
\{"score": <int between 1 and 10>\}\\
\\
\textbf{User message:}\\
\\
\textbf{\#\#\# User's Request}\\
\\
\{user\_request\}\\
\\
\textbf{\#\#\# AI Assistant's Response}\\
\\
\{response\}\\

\bottomrule
\end{tabularx}
}
\label{prompt:pfr-judge}
\end{table*}

%% file: prompts/PLR.tex
\begin{table*}[h]
\centering
\caption{LLM-as-a-judge prompt used to evaluate preference leakage for Suppress preferences.}
\scalebox{0.77}{
\begin{tabularx}{\textwidth}{X}
\toprule
\textbf{Preference Leakage Judge (PLR)} \\
\midrule

\textbf{\#\# Your Objective}\\
\\
You are a critical and meticulous evaluator. You will be presented with a user's request to an AI assistant and the AI's response to the user. Your task is to evaluate whether the AI assistant's response \textbf{inappropriately reflected} preferences that should have been \textbf{suppressed} in the given context. To help you evaluate, you will also be provided with a \textbf{leakage checklist} for each suppressed preference that decomposes it into specific questions about whether signs of that preference appear in the response.\\
\\
\textbf{\#\#\# Suppressed Preferences and Leakage Checklists}\\
\\
\textbf{Suppressed Preference 1}: "\{preference\_1\}"\\
\textbf{Leakage Checklist}:\\
\hspace{3mm}- \{checklist\_item\_1\}\\
\hspace{3mm}- \{checklist\_item\_2\}\\
\\
$\cdots$\\
\\
\textbf{\#\# Instructions}\\
\\
You should write down your analysis and assessment on how much each suppressed preference leaks into the AI assistant's response. You should follow these considerations:\\
\\
- Walk through each checklist item and note any "signs of leakage" or "correct suppression" in the response regarding that item.\\
- For each checklist item, you should consider whether the suppressed preference is inappropriately reflected (leakage) or correctly absent.\\
- Avoid considering aspects that are not included in the checklist. Focus only on the leakage checklist. Ensure that your persona profile does not influence your evaluation.\\
- You should then return a score in the range of 1--10, where 1 means no leakage (all suppressed preferences are correctly absent) and 10 means heavy leakage (suppressed preferences are strongly reflected throughout the response).\\
\\
Ensure that you follow the format given below. Avoid adding additional content that is not included in the format below.\\
\\
---\\
\\
\textbf{\#\# Output Format}\\
\\
\textbf{\#\#\# Evaluation of AI Assistant's Response}\\
\\
\textbf{Suppressed Preference 1: \{preference\_1\}}\\
1. \textbf{\{checklist\_item\_1\}}: <Identify any signs of leakage or correct suppression regarding this checklist item>\\
2. \textbf{\{checklist\_item\_2\}}: <Identify any signs of leakage or correct suppression regarding this checklist item>\\
\\
$\cdots$\\
\\
\textbf{\#\#\# Evaluation Score}\\
\\
\{"score": <int between 1 and 10>\}\\
\\
\textbf{User message:}\\
\\
\textbf{\#\#\# User's Request}\\
\\
\{user\_request\}\\
\\
\textbf{\#\#\# AI Assistant's Response}\\
\\
\{response\}\\

\bottomrule
\end{tabularx}
}
\label{prompt:plr-judge}
\end{table*}

%% file: prompts/reasoning_extr.tex
\begin{table*}[h]
\centering
\caption{Reasoning-label extraction prompt used to evaluate preference decisions expressed in the model's reasoning trace.}
\scalebox{0.9}{
\begin{tabularx}{\textwidth}{X}
\toprule
\textbf{Reasoning Label Extraction} \\
\midrule
You are an expert evaluator analyzing a language model's preference personalization behavior.\\
\\
You will receive:\\
1. A numbered list of user preferences (each may be "Apply" or "Suppress" in the ground truth)\\
2. The model's chain-of-thought REASONING only (do NOT consider any final response)\\
\\
Your job is to extract reasoning\_labels --- what the model's REASONING indicates for each preference:\\
\hspace{3mm}"apply" --- reasoning explicitly states this preference should be followed\\
\hspace{3mm}"suppress" --- reasoning explicitly states this preference should be avoided/ignored\\
\hspace{3mm}"none" --- preference is not addressed or is ambiguous in the reasoning\\
\\
Return ONLY valid JSON (no markdown fences) in exactly this structure:\\
\{\\
\hspace{3mm}"reasoning\_labels": \{"Preference 1": "Apply", "Preference 2": "Suppress", "Preference 3": "None", ...\}\\
\}\\
\bottomrule
\end{tabularx}
}
\label{prompt:reasoning-label-extraction}
\end{table*}

%% file: prompts/generation_Extr.tex
\begin{table*}[h]
\centering
\caption{Generation-label extraction prompt used to evaluate whether each user preference is reflected in the model's final response.}
\scalebox{0.9}{
\begin{tabularx}{\textwidth}{X}
\toprule
\textbf{Generation Label Extraction} \\
\midrule
You are an expert evaluator analyzing a language model's preference personalization behavior.\\
\\
You will receive:\\
1. A numbered list of user preferences (each may be "Apply" or "Suppress" in the ground truth)\\
2. The model's final RESPONSE only (do NOT consider any reasoning or thinking text)\\
\\
Your job is to extract generation\_labels --- what the final RESPONSE actually demonstrates for each preference:\\
\hspace{3mm}"apply" --- the preference is reflected in the response (the model incorporated it)\\
\hspace{3mm}"suppress" --- the preference is not reflected in the response (the model ignored/omitted it)\\
\\
Return ONLY valid JSON (no markdown fences) in exactly this structure:\\
\{\\
\hspace{3mm}"generation\_labels": \{"Preference 1": "Apply", "Preference 2": "Suppress", "Preference 3": "Apply", ...\}\\
\}\\
\bottomrule
\end{tabularx}
}
\label{prompt:generation-label-extraction}
\end{table*}

%% file: prompts/decide_only.tex
\begin{table*}[h]
\centering
\caption{Direct Decision (Step 1) for conditionally applying or suppressing user preferences based on context in ABIDE.}
\scalebox{0.9}{ 
\begin{tabularx}{\textwidth}{X}
\toprule
\textbf{Direct Decision (Step 1) in ABIDE} \\
\midrule
\textbf{Step 1:} Decide whether user preferences should be applied or suppressed in a given context.\\
For each preference, output exactly "Apply" or "Suppress".\\
Rules:\\
"Apply" means the preference is relevant and appropriate for the current context.\\
"Suppress" means the preference is irrelevant, inappropriate, or would conflict with the context's norms or the user's actual intent.\\
Present your analysis under the header:\\
Step 1 output:\\
\{step1\_format\}\\
\bottomrule
\end{tabularx}
}
\label{prompt:Decide}
\end{table*}

%% file: prompts/neutral.tex
\begin{table*}[h]
\centering
\caption{Neutral 2-step prompt template for conditionally applying or suppressing user preferences based on context.}
\scalebox{0.9}{ 
\begin{tabularx}{\textwidth}{X}
\toprule
\textbf{Neutral 2-step prompt} \\
\midrule
\textbf{Step 1:} Decide whether user preferences should be applied or suppressed in a given context.\\
For each preference, output exactly "Apply" or "Suppress".\\
Rules:\\
"Apply" means the preference is relevant and appropriate for the current context.\\
"Suppress" means the preference is irrelevant, inappropriate, or would conflict with the context's norms or the user's actual intent.\\
Present your analysis under the header:\\
Step 1 output:\\
\{step1\_format\}\\
\\
\textbf{Step 2:} Output the fixed sentence: "Task completed." \\
Your Step 2 output must contain only the content addressing the user's request do not include your reasoning or preference classifications.\\
Present your response under the header:\\
Step 2 output: \\
\bottomrule
\end{tabularx}
}
\label{prompt:N}
\end{table*}

%% file: prompts/G2.tex
\begin{table*}[h]
\centering
\caption{Prompt configuration for the Simplified Generation Objective (G2), which removes explicit ``Apply'' and ``Suppress'' terminology from the generation instruction while preserving the preference discrimination step.}
\scalebox{0.9}{ 
\begin{tabularx}{\textwidth}{X}
\toprule
\textbf{Simplified Generation Objective (G2)} \\
\midrule
\textbf{Step 1:} Decide whether user preferences should be applied or suppressed in a given context.\\
For each preference, output exactly "Apply" or "Suppress".\\
Rules:\\
"Apply" means the preference is relevant and appropriate for the current context.\\
"Suppress" means the preference is irrelevant, inappropriate, or would conflict with the context's norms or the user's actual intent.\\
Present your analysis under the header:\\
Step 1 output:\\
\{step1\_format\}\\
\\
\textbf{Step 2:} Write a response that fully addresses the user's request, consistent with your Step 1 analysis.\\
Your Step 2 output must contain only the content addressing the user's request do not include your reasoning or preference classifications.\\
Present your response under the header:\\
Step 2 output:\\
\bottomrule
\end{tabularx}
}
\label{prompt:G2}
\end{table*}

%% file: prompts/N2.tex
\begin{table*}[h]
\centering
\caption{Prompt configuration for the Arithmetic Neutral Objective (N2). The objective replaces the fixed-sentence output in the neutral baseline with an arithmetic computation in Step 2, while keeping the preference discrimination task in Step 1 unchanged.}
\scalebox{0.9}{ 
\begin{tabularx}{\textwidth}{X}
\toprule
\textbf{Arithmetic Neutral Objective (N2)} \\
\midrule
\textbf{Step 1:} Decide whether user preferences should be applied or suppressed in a given context.\\
For each preference, output exactly "Apply" or "Suppress".\\
Rules:\\
"Apply" means the preference is relevant and appropriate for the current context.\\
"Suppress" means the preference is irrelevant, inappropriate, or would conflict with the context's norms or the user's actual intent.\\
Present your analysis under the header:\\
Step 1 output:\\
\{step1\_format\}\\
\\
\textbf{Step 2:} Compute the value of the arithmetic expression $3+3+9.$\\
Your Step 2 output must contain only the numeric result do not include your reasoning or preference classifications.\\
Present your response under the header:\\
Step 2 output: \\
\bottomrule
\end{tabularx}
}
\label{prompt:N2}
\end{table*}